\documentclass[runningheads]{llncs}

\usepackage{eccv}

\usepackage{eccvabbrv}

\usepackage{caption}
\usepackage{subcaption}
\usepackage{tabularx, booktabs} 
\usepackage{rotating} 
\usepackage{enumitem}
\usepackage{algorithm}
\usepackage{algpseudocode}
\usepackage{multirow}
\usepackage{array}
\usepackage{colortbl}
\usepackage{color}
\algnewcommand\INPUT{\item[\textbf{Input:}]}%
\algnewcommand\OUTPUT{\item[\textbf{Output:}]}%
\usepackage[normalem]{ulem}
\usepackage{diagbox}
\usepackage{adjustbox}
\usepackage{pdfpages}

\usepackage[accsupp]{axessibility}  

\usepackage{hyperref}

\usepackage{orcidlink}

\begin{document}

\title{FYI: Flip Your Images for Dataset Distillation} 


\author{Byunggwan Son\orcidlink{0009-0000-7344-0355} \and
Youngmin Oh\orcidlink{0009-0006-5568-2127} \and \\
Donghyeon Baek\orcidlink{0009-0003-2470-1469} \and
Bumsub Ham\thanks{Corresponding author.}\orcidlink{0000-0002-3443-8161} \\ \url{https://cvlab.yonsei.ac.kr/projects/FYI}}

\authorrunning{B.~Son et al.}

\institute{Yonsei University}

\maketitle

\begin{abstract}
    Dataset distillation synthesizes a small set of images from a large-scale real dataset such that synthetic and real images share similar behavioral properties~(e.g,~distributions of gradients or features) during a training process. Through extensive analyses on current methods and real datasets, together with empirical observations, we provide in this paper two important things to share for dataset distillation. First, object parts that appear on one side of a real image are highly likely to appear on the opposite side of another image within a dataset, which we call the bilateral equivalence. Second, the bilateral equivalence enforces synthetic images to duplicate discriminative parts of objects on both the left and right sides of the images, limiting the recognition of subtle differences between objects. To address this problem, we introduce a surprisingly simple yet effective technique for dataset distillation, dubbed FYI, that enables distilling rich semantics of real images into synthetic ones. To this end, FYI embeds a horizontal flipping technique into distillation processes, mitigating the influence of the bilateral equivalence, while capturing more details of objects. Experiments on CIFAR-10/100, Tiny-ImageNet, and ImageNet demonstrate that FYI can be seamlessly integrated into several state-of-the-art methods, without modifying training objectives and network architectures, and it improves the performance remarkably.
\end{abstract}

\section{Introduction}
\label{sec:intro}
\begin{figure}[t]
    \captionsetup[subfigure]{labelformat=empty}
    \tiny
    \centering
  
    \begin{subfigure}{.18\linewidth}
      \centering
      \includegraphics[width=\textwidth]{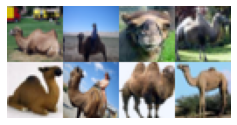}\vspace{0.1cm}
    \end{subfigure}
    \begin{subfigure}{.18\linewidth}
      \centering
      \includegraphics[width=\textwidth]{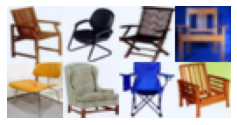}\vspace{0.1cm}
    \end{subfigure}
    \begin{subfigure}{.18\linewidth}
      \centering
      \includegraphics[width=\textwidth]{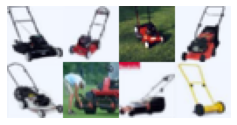}\vspace{0.1cm}
    \end{subfigure}
    \begin{minipage}{\linewidth}
      \centering
      \small{(a) Real images}
    \end{minipage}
    \hfil
  
    \begin{subfigure}{.18\linewidth}
      \centering
      \includegraphics[width=\textwidth]{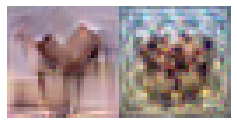}\vspace{0.1cm}
    \end{subfigure}
    \begin{subfigure}{.18\linewidth}
      \centering
      \includegraphics[width=\textwidth]{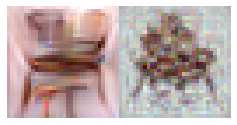}\vspace{0.1cm}
    \end{subfigure}
    \begin{subfigure}{.18\linewidth}
      \centering
      \includegraphics[width=\textwidth]{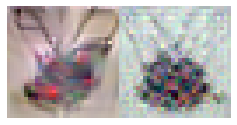}\vspace{0.1cm}
    \end{subfigure}
    \begin{minipage}{\linewidth}
      \centering
      \small{(b) MTT~\cite{mtt} and DSA~\cite{dsa}}
    \end{minipage}
    \hfil
  
    \begin{subfigure}{.18\linewidth}
      \centering
      \includegraphics[width=\textwidth]{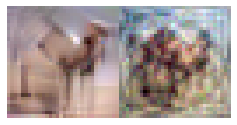}\vspace{0.1cm}
    \end{subfigure}
    \begin{subfigure}{.18\linewidth}
      \centering
      \includegraphics[width=\textwidth]{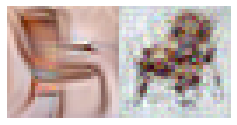}\vspace{0.1cm}
    \end{subfigure}
    \begin{subfigure}{.18\linewidth}
      \centering
      \includegraphics[width=\textwidth]{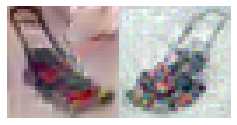}\vspace{0.1cm}
    \end{subfigure}
    \begin{minipage}{\linewidth}
      \centering
      \small{(c) MTT+FYI and DSA+FYI}
    \end{minipage}
    \hfil
  
    \caption{\small{Comparisons of existing dataset distillation methods and our approach with the 1 IPC setting on CIFAR-100~\cite{cifar}: Camel, chair, and lawn mower classes. (a) Objects in natural images are oriented diversely, and (b) current dataset distillation methods~((left)~MTT~\cite{mtt} and (right)~DSA~\cite{dsa}) synthesize symmetric images with repeated patterns in the left and right halves, neglecting fine-grained details of objects. (c) Applying FYI to MTT and DSA avoids this problem, while capturing the fine-grained details.}}
    \vspace{-.7cm}
    \label{fig:teaser}
  \end{figure}
Training neural networks~\cite{resnet,vgg,vit,clip} with large-scale datasets~\cite{imagenet,jft300m,imagenet21k} is computationally expensive, and also requires lots of memory for storing training samples. Dataset distillation~\cite{DD} addresses this problem by condensing entire training samples into a small set of synthetic images and training networks with the synthetic ones. This facilitates many applications, including continual learning~\cite{icarl,der,co2l}, neural architecture search (NAS)~\cite{nasnet,enas,darts,spos}, and federated learning~\cite{fedavg,fedprox,li2019convergence}. For example, it is important in NAS to predict the performance of an arbitrary architecture efficiently. We can use synthetic images obtained from dataset distillation methods as proxies for original training samples. The networks trained with the synthetic images can then be used to predict the performance, instead of training networks with the original samples.

The seminal work of~\cite{DD} formulates the dataset distillation task as a bi-level optimization problem. Specifically, it trains neural networks with synthetic images, while optimizing the synthetic images with the trained networks alternately. This approach, however, requires numerous updates to train the networks using synthetic images~\cite{rtp}. Recent works avoid the iterative updates by approximating the training process with ridge regression using the neural tangent kernel (NTK)~\cite{kip1,kip2} or exploiting surrogate objectives encouraging real and synthetic images to have similar properties~(\eg, gradients~\cite{dc,dsa,dcc,idc,dream}, network trajectories~\cite{mtt,ftd}, or feature distributions~\cite{cafe,dm,datadam}) during the training process. Although these methods achieve better results in terms of efficiency and accuracy, we have observed that they produce similar patterns in the left and right halves across a synthetic dataset, failing to distill various semantics of real datasets into synthetic ones. For example, \cref{fig:teaser}(b) shows a single image per class (IPC) synthesized using current dataset distillation methods~\cite{mtt,dsa}. We can see that the synthetic images are highly likely to be symmetric. Particularly, both halves of the synthetic images contain discriminative parts of objects (\eg, the back support of a chair), which rather prevent the synthetic images from capturing fine-grained details. The reason behind this is that similar object parts are present on the left and right sides equivalently in a real dataset~(\cref{fig:teaser}(a)); a phenomenon we call the bilateral equivalence. One potential solution to consider the bilateral equivalence of real datasets is to align images in the dataset before applying dataset distillation methods. However, aligning several images is nontrivial, especially when there are many objects in the images and/or objects are occluded.

In this paper, we propose a surprisingly simple yet effective method, dubbed FYI, that embeds a horizontal flipping technique into a dataset distillation process. Exploiting synthetic images with horizontally flipped counterparts reduces duplicated patterns remarkably, preventing a discriminative part synthesized on one side of a specific image from being duplicated on the other side of the image, as well as on any side of other images. For example, the lawn mower in \cref{fig:teaser}(c) contains fine-grained details with distinguishable front and back parts, compared to that in \cref{fig:teaser}(b). FYI can easily be integrated with existing dataset distillation methods to boost the performance, and it encourages them to transfer rich semantics from real to synthetic images, providing more clues when training networks with synthetic images. Extensive experiments on standard benchmarks~\cite{cifar,tiny,mtt,insub} demonstrate that FYI improves the performance of existing dataset distillation methods significantly, especially for fine-grained classification~\cite{mtt}. We summarize our contributions in the following:

\begin{itemize}
  \item[$\bullet$] We provide in-depth analyses on the bilateral equivalence for dataset distillation, and show that existing methods fail to encode diverse semantics of objects.
  \item[$\bullet$] In order to consider the bilateral equivalence for dataset distillation, we introduce a generic approach, dubbed FYI, that can be applied to any dataset distillation methods to prevent parts of objects synthesized within one side of an image from being duplicated on the other side of the image and on different images.
  \item[$\bullet$] We demonstrate the effectiveness of FYI through comprehensive experiments across various combinations of dataset distillation methods~\cite{dc,dsa,dm,mtt}, datasets~\cite{cifar,tiny,mtt,insub}, and compression ratios.
\end{itemize}

\section{Related work}
\label{sec:related_work}
Dataset distillation condenses a set of natural images into a few synthetic ones, which can be categorized into two groups, regression-based and matching-based approaches. The first approach synthesizes images using a kernel ridge regression method.  Specifically, it tries to regress real images from synthetic ones in a feature space. For example, KIP~\cite{kip1,kip2} performs regression using NTKs~\cite{ntk} that represent training dynamics of neural networks~\cite{ntkapprox}. FRePo~\cite{frepo} instead uses convolutional features to avoid the expensive calculation for NTKs. Kernel ridge regression exploits all synthetic images at each training step, which is computationally expensive, and thus it would be not adequate for large-scale datasets~\cite{tesla}. To overcome the scalability issue, the second approach optimizes synthetic images, such that real and synthetic images share similar behavioral properties during a training process. DC~\cite{dc} enforces synthetic and real images to have similar gradients at every training step. Extending the single-step approach of DC~\cite{dc}, MTT~\cite{mtt} proposes to imitate long-range trajectories of optimization steps for real images, in order for synthetic images to better mimic the training dynamics of real images. The works of~\cite{cafe,dm,m3d} enforce real and synthetic images to have similar feature statistics, by minimizing the maximum mean discrepancy~\cite{mmd} between intermediate features of these images, which is more efficient compared to other methods~\cite{idm}. We have observed that all the aforementioned methods encode similar semantics repeatedly on one side of an image and the other side of the same or a different image, regardless of the training objectives, which distracts from distilling rich semantics into the synthetic images.

Other approaches attempt to adjust real and/or synthetic images before applying dataset distillation methods. DREAM~\cite{dream} uses the K-means clustering technique~\cite{kmeans} to sample real images representing entire training samples. Although this method accelerates the training speed, and shows a satisfactory distillation performance, the representative images still contain objects with diverse orientations, providing the bilateral equivalence. DSA~\cite{dsa} proposes to apply a data augmentation technique to both real and synthetic images in order to consider the effect of the augmentation for training networks with synthetic images. Our approach also exploits a data augmentation technique (\ie,~horizontal flipping), but differs in that we focus on distilling rich semantics from real images into synthetic ones, rather than learning how the real images respond to the augmentation technique for dataset distillation. Recently, the works of~\cite{idc,idm,rtp,haba,sparse} propose to parameterize synthetic images in order to encode rich semantics from a set of natural images more efficiently within limited storage. Specifically, HaBa~\cite{haba} feeds class-specific latent codes into lightweight networks to form synthetic images. IDC~\cite{idc} synthesizes low-resolution images which are then up-sampled using bilinear interpolation, assuming that nearby pixels are similar. Our approach also transfers rich semantics from real to synthetic images, but for the purpose of mitigating the influence of the bilateral equivalence, which has not been addressed by the previous methods.

\section{Method}
In this section, we describe dataset distillation briefly (\cref{sec:preliminary}) and analyze the bilateral equivalence~(\cref{sec:problem}). We then present a detailed description of our approach~(\cref{sec:fyi}).

\subsection{Problem statement}
\label{sec:preliminary}
Let us denote by $\mathcal{T}_c$ and $\mathcal{S}_c$ sets of real and synthetic images for the class $c$, respectively, defined as follows:
\begin{equation}
    \small
    \begin{split}
        \mathcal{T}_{c} &= 
        \{ t_i \mid i = 1,\dots,N_c \}, \\
        \mathcal{S}_{c} &= \{ s_j \mid j = 1,\dots,M_c \},
    \end{split}
\end{equation}
where $t_i$ and $s_j$ indicate real and synthetic images, respectively. Note that the number of real images is much larger than that of synthetic images~(\ie,~$N_c \gg M_c$). The goal of dataset distillation methods is to estimate a small set of synthetic images such that networks trained on the set provide results similar to those trained on the real dataset in terms of accuracy. To this end, current methods~\cite{dc,dm,mtt} imitate the training process of real images. Specifically, they define a distance metric $D_\theta$ quantifying the difference between two datasets in terms of gradients~\cite{dc,mtt} or convolutional features~\cite{dm} for the network parameterized by $\theta$, and minimize an objective function over various networks as follows:
\begin{equation}
    \small
    \label{eq:1}
    \mathcal{L} = \mathbb{E}_{\theta\sim P_\theta}\Bigl[\sum_c D_\theta(\mathcal{T}_c, \mathcal{S}_c)\Bigr],
\end{equation}
where we denote by $P_\theta$ a distribution of network parameters. For example, DM~\cite{dm} exploits the following distance metric\footnote{Here we mainly describe our approach based on DM. Detailed descriptions for other methods, including DC~\cite{dc} and MTT~\cite{mtt}, can be found in the supplementary material.}:
\begin{equation}
    \small
    \label{eq:2}
    D_\theta(\mathcal{T}_c, \mathcal{S}_c)= \Bigl\Vert\frac{1}{N_c}\sum_{i}C_{\theta}(t_i) - \frac{1}{M_c}\sum_{j}C_{\theta}(s_j)\Bigr\Vert^2,
\end{equation}
where $\Vert\cdot\Vert$ is the Euclidean distance, and $C_{\theta}$ computes convolutional features using a network parameterized by $\theta$. The synthetic images for the class $c$ are then optimized as follows:
\begin{equation}
    \small
    \label{eq:3}
    \mathcal{S}_c \leftarrow \mathcal{S}_c - \eta \frac{\partial D_\theta(\mathcal{T}_c, \mathcal{S}_c)}{\partial \mathcal{S}_c},
\end{equation}
where $\eta$ is a learning rate. In this way, DM encourages synthetic images to imitate an average feature of real images. However, we have found that current methods~\cite{dc,dm,mtt} fail to capture fine-grained details of real images, distilling a few discriminative patterns into the synthetic images only. In the following, we describe this problem in detail.
\begin{figure}[t]
    \captionsetup{font={small}}
    \begin{center}
       \includegraphics[width=0.18\linewidth]{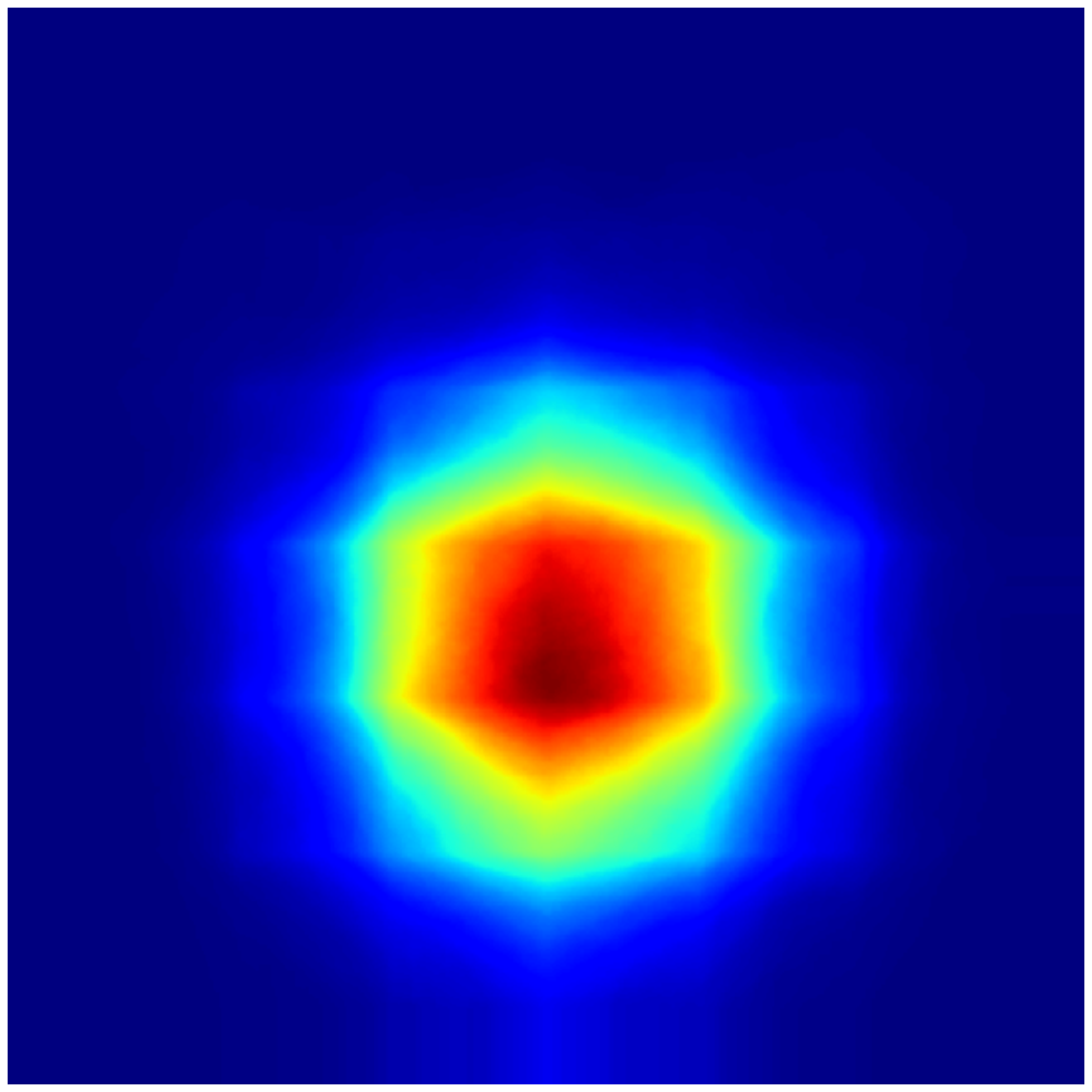}\hspace{0.01\linewidth}%
       \includegraphics[width=0.18\linewidth]{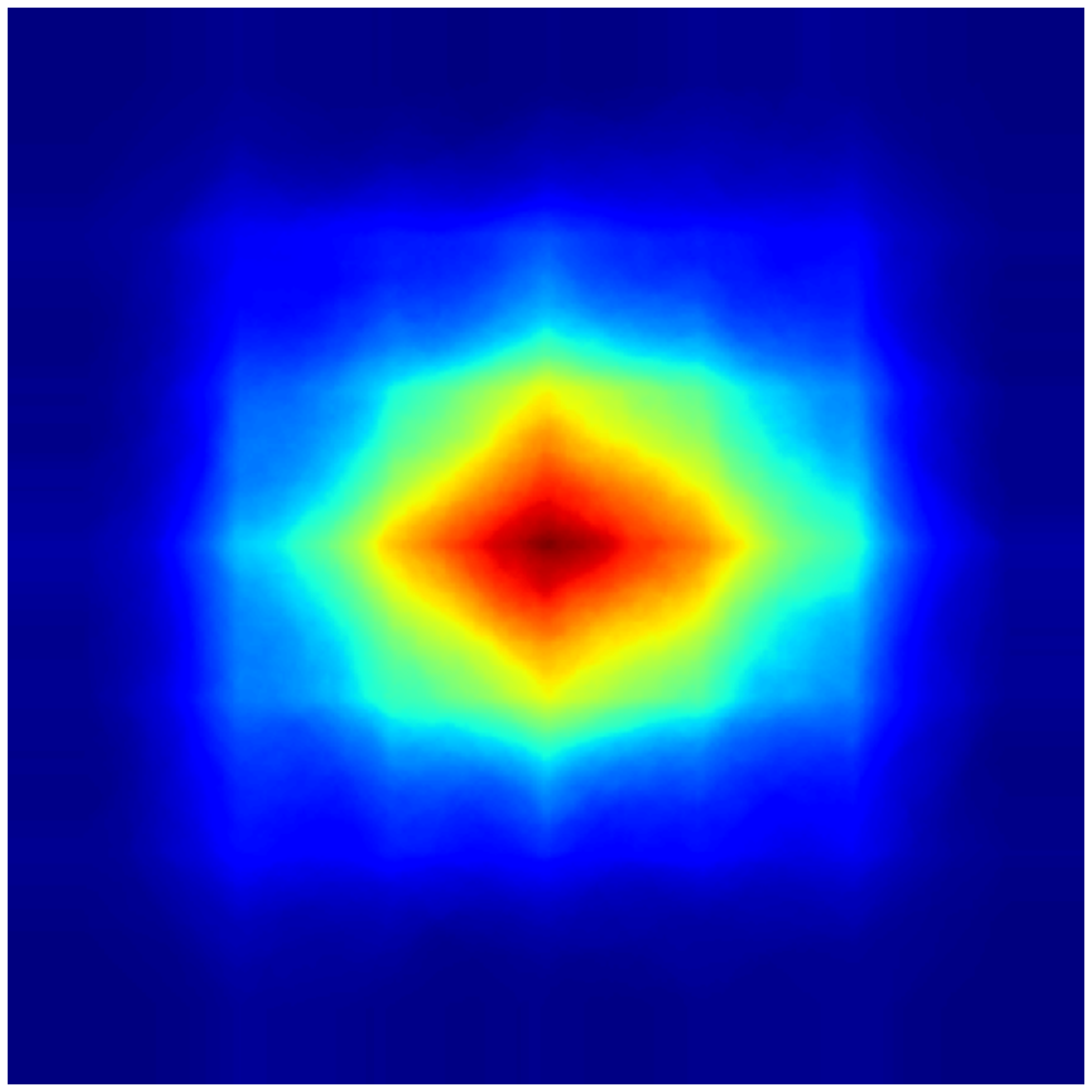}\hspace{0.01\linewidth}%
       \includegraphics[width=0.18\linewidth]{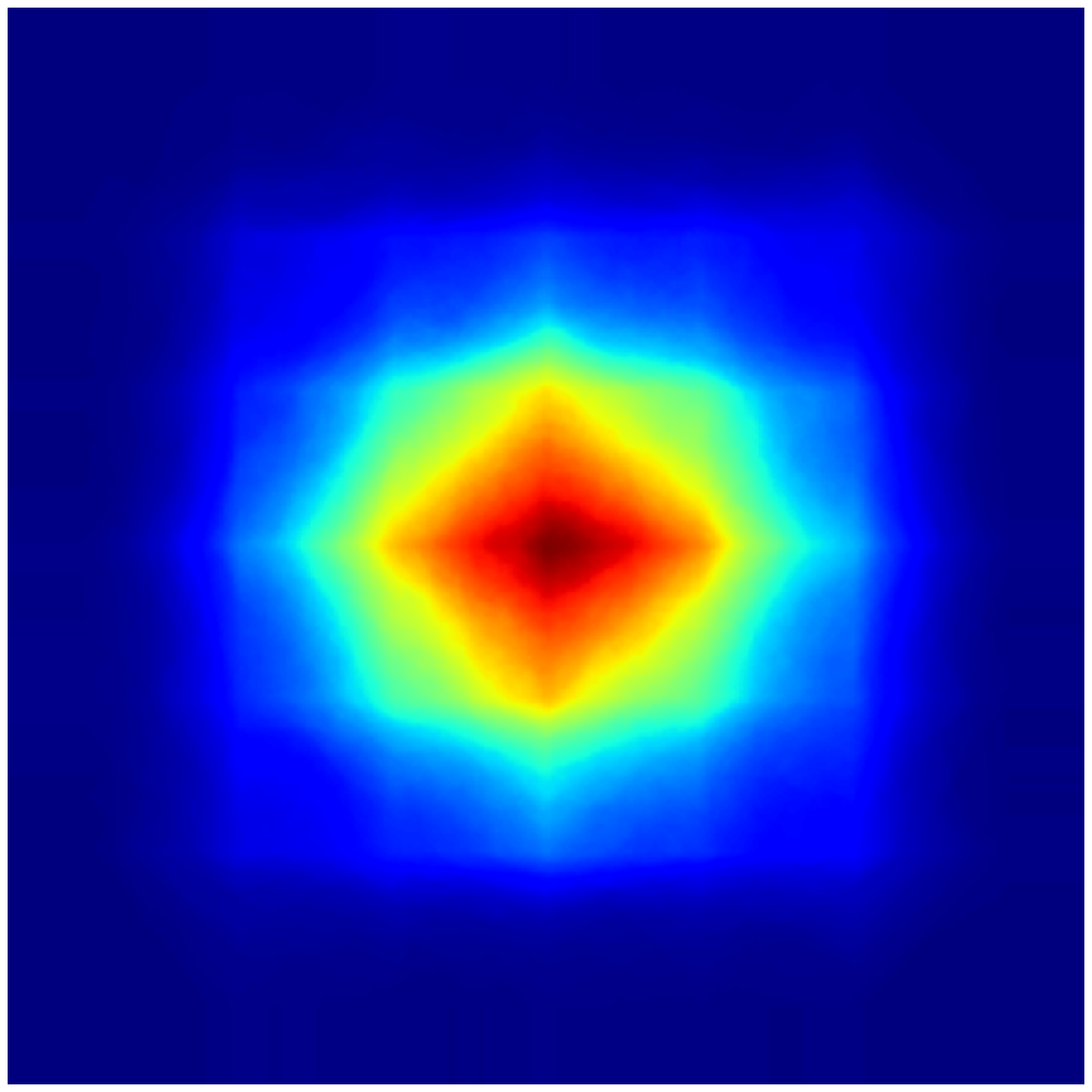}\hspace{0.01\linewidth}%
       \includegraphics[width=0.18\linewidth]{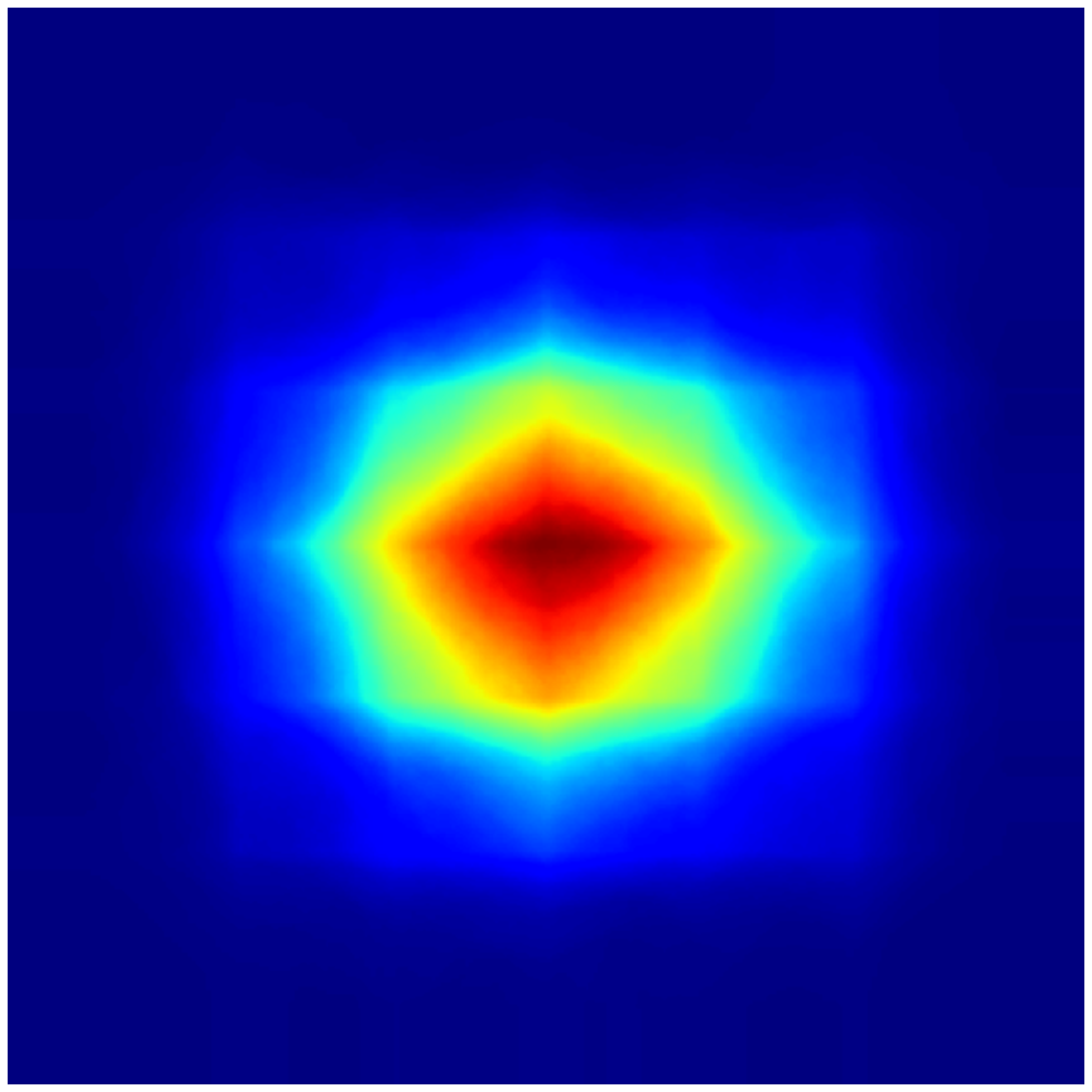}\hspace{0.01\linewidth}%
       \includegraphics[width=0.18\linewidth]{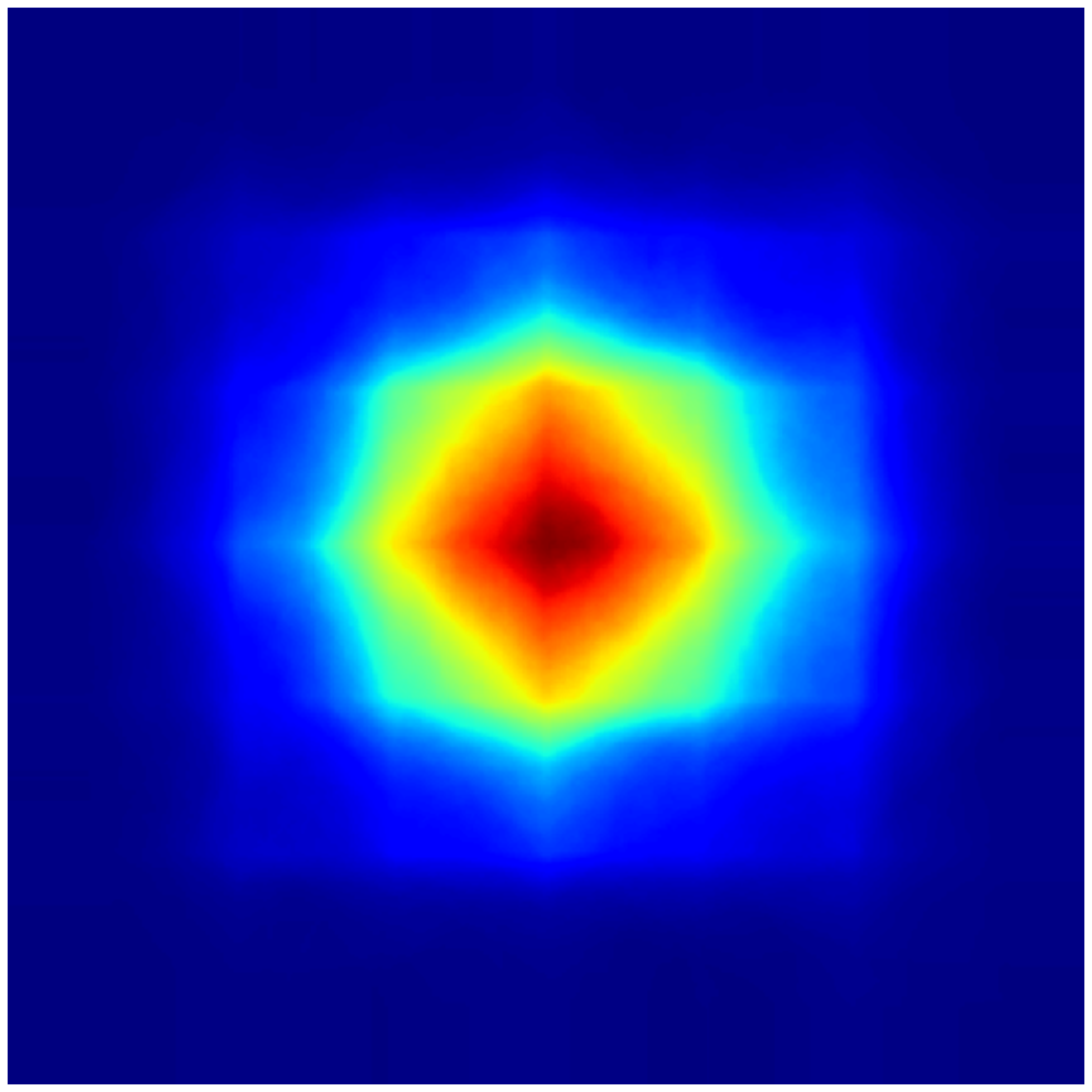}
    \end{center}
    \vspace{-1.0cm}
    \caption{Distributions of discriminative object parts for the class of (from left to right) tench, goldfish, white shark, tiger shark, and hammerhead, on  ImageNet~\cite{imagenet}. We count how many times each pixel belongs to the top-10\% of attention values obtained from class activation maps~\cite{cam} using a pre-trained ResNet-18~\cite{resnet}. Red: high, Blue: low.}
    \label{fig:heatmap}
    \vspace{-.5cm}
\end{figure}

\subsection{The bilateral equivalence}
\label{sec:problem}
\begin{figure}[t]
     \captionsetup{font={small}}
     \begin{center}
        \begin{subfigure}{\linewidth}
           \centering
           \captionsetup{justification=centering}
           \hspace{0.5cm}\subcaptionbox{DC~\cite{dc}}{
           \hspace{-0.5cm}\includegraphics[width=0.23\linewidth]{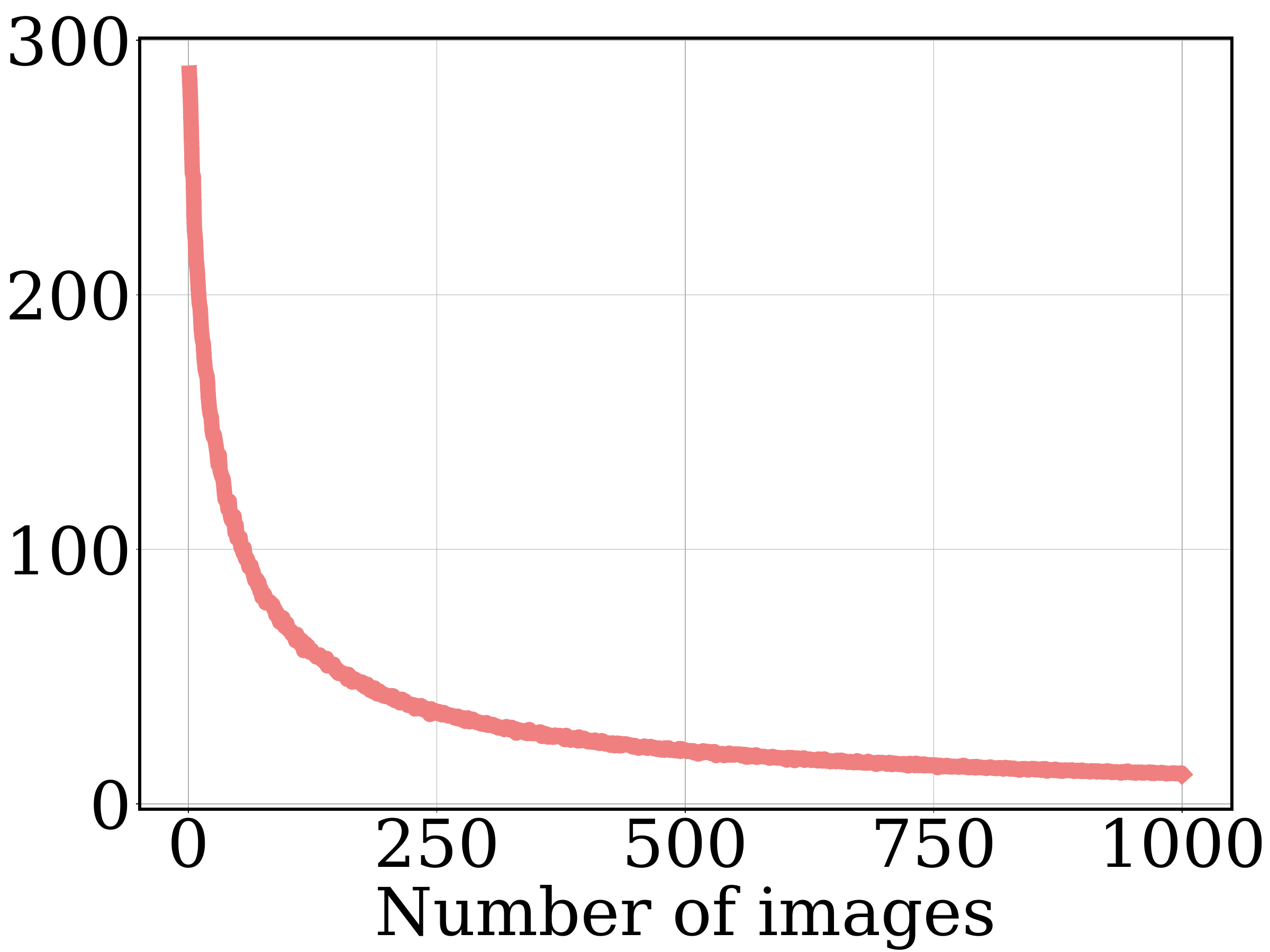}}
           \hspace{0.5cm}\subcaptionbox{DSA~\cite{dsa}}{
           \hspace{-0.5cm}\includegraphics[width=0.23\linewidth]{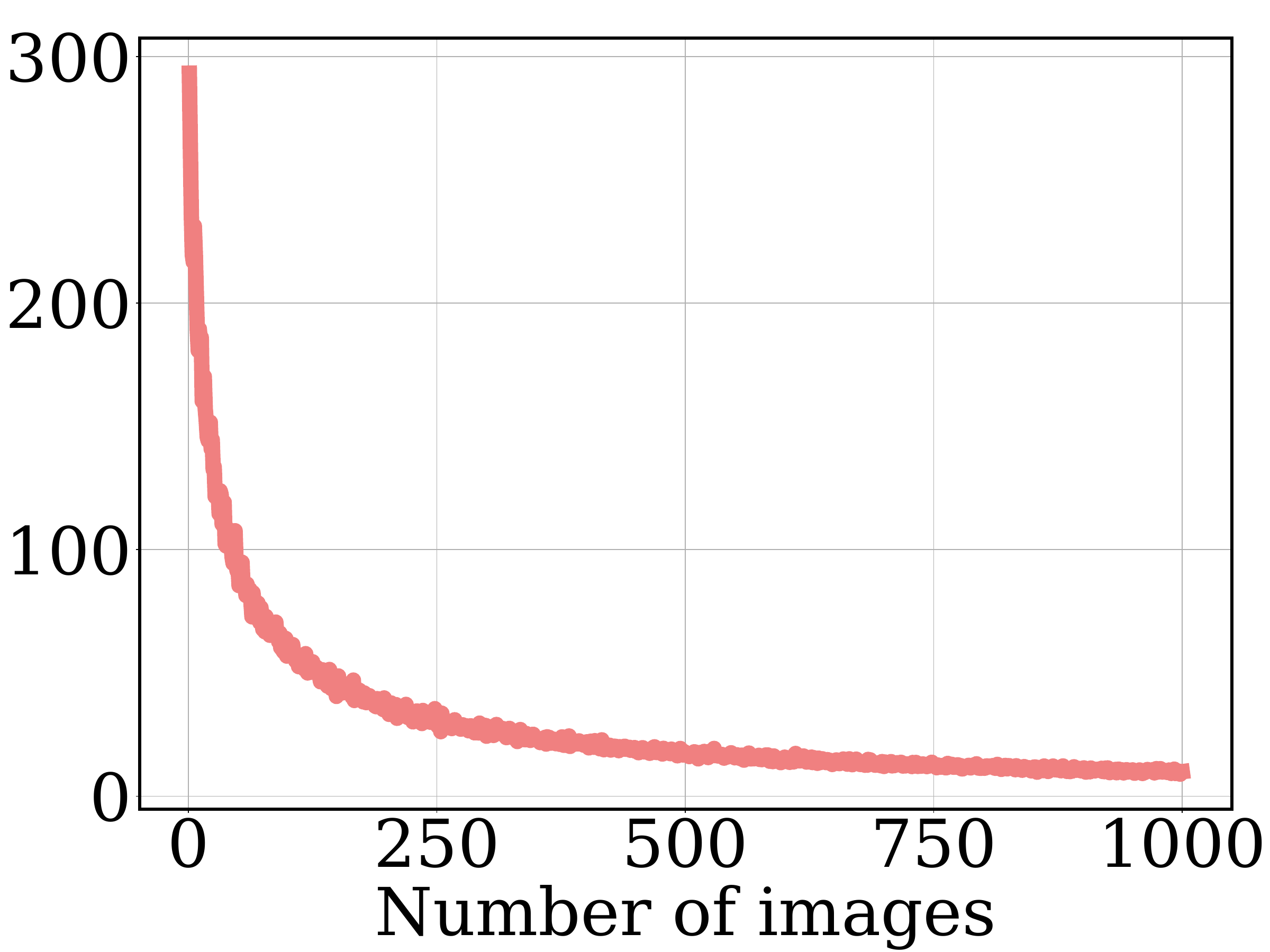}}
           \hspace{0.5cm}\subcaptionbox{DM~\cite{dm}}{
            \hspace{-0.5cm}\includegraphics[width=0.23\linewidth]{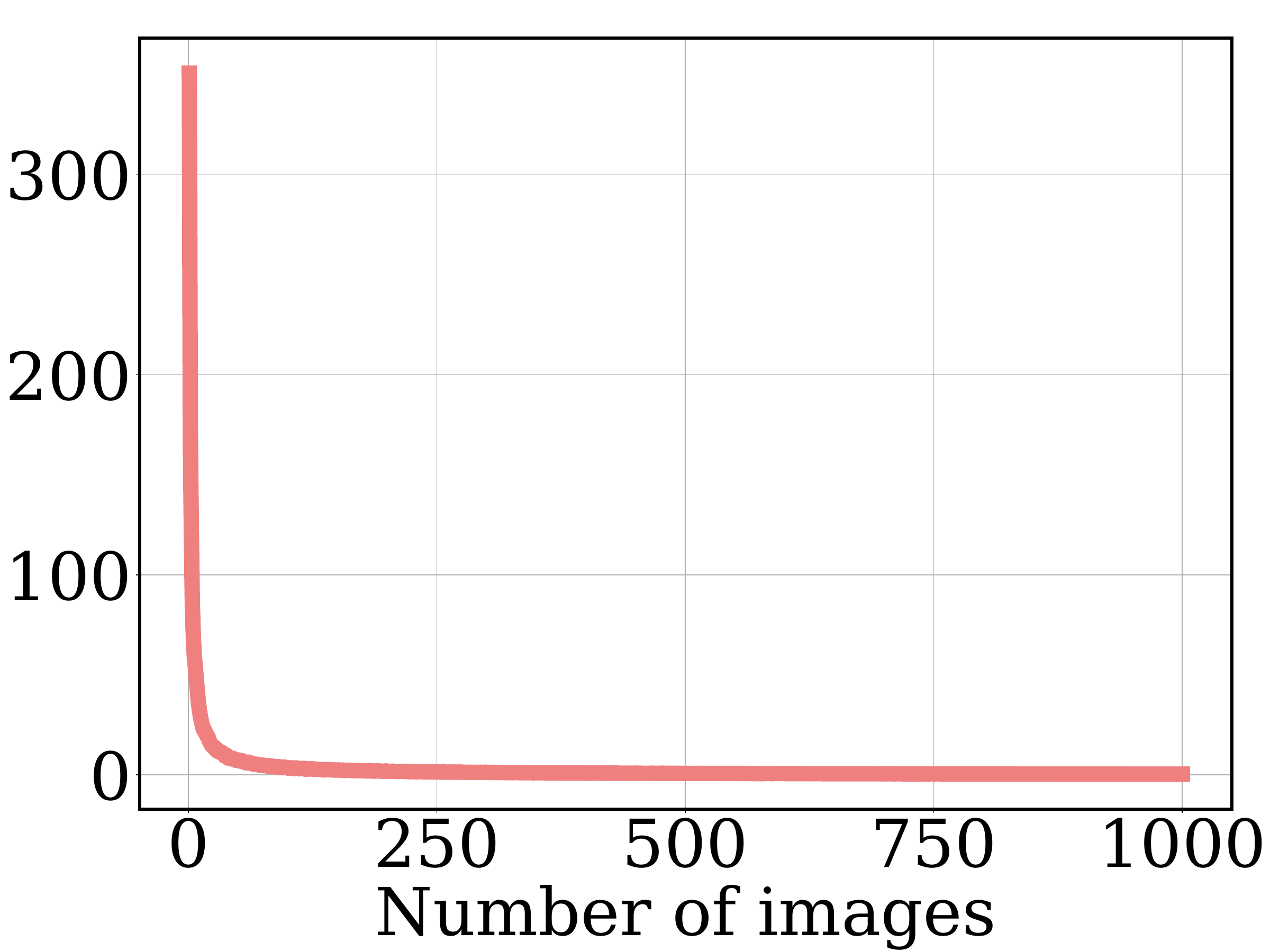}}
        \end{subfigure}
     \end{center}
     \vspace{-0.5cm}
        \caption{The bilateral equivalence of a real-world dataset. We compute the unequalness score with a set of image(s) for each object class, where we randomly sample the images from CIFAR-10~\cite{cifar}, using DC~\cite{dc}, DSA~\cite{dsa} and DM~\cite{dm} as distance metrics in~\cref{eq:4}, and show the scores averaged over the classes.}
     \label{fig:realscore}
\end{figure}
\begin{figure}[t]
    \begin{center}
       \includegraphics[width=.7\linewidth]{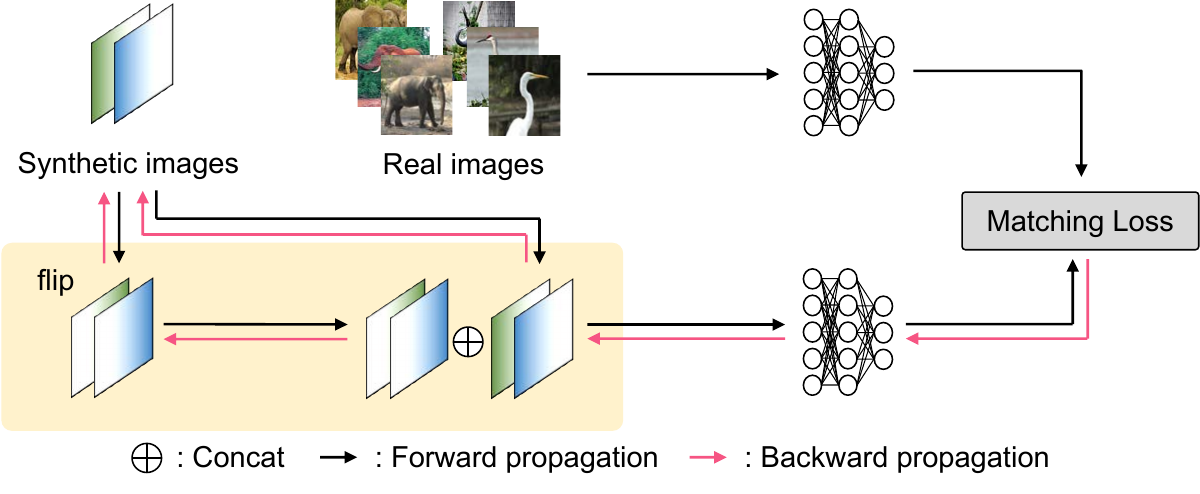}
    \end{center}
    \vspace{-.5cm}
    \caption{FYI augments synthetic images with the flipped counterparts to avoid the influence of the bilateral equivalence for dataset distillation.}
    \label{fig:overview}
    \vspace{-.5cm}
\end{figure}
It is unlikely that a specific part of objects consistently appears on either the left or right halves of natural images. That is, particular patterns (\emph{e.g.}, a head of an animal) could be on the left or right side of images with an equal possibility in a balanced manner, which we call the bilateral equivalence. We show in \cref{fig:heatmap} distributions of positions for the top-10\% of the attention values obtained using class activation maps~\cite{cam} for real images of different object classes. We can observe that the distributions are highly symmetric, since discriminative parts of objects tend to distribute equally to the left and right sides. To concretely analyze the bilateral equivalence, we define an unequalness score of an arbitrary set of images $\mathcal{R}$ using the distance metric in~\cref{eq:1} as follows:
\begin{equation}
    \small
    \label{eq:4}
    \text{Score}(\mathcal{R}) = D_\theta\left(\mathcal{R}, \text{Flip}(\mathcal{R})\right),
\end{equation}
where $\text{Flip}$ is a function that flips an image or all the images in a set horizontally. Note that the score is zero if a set $\mathcal{R}$ is flip-invariant, \emph{i.e.}, $\text{Flip}(\mathcal{R}) = \mathcal{R}$. A set is flip-invariant if the set contains a flipped counterpart for every image, \emph{i.e.}, $\forall r \in \mathcal{R}$, $\text{Flip}(r) \in \mathcal{R}$. Thus, the unequalness score approaches to zero, as more similar patterns appear evenly on different sides of images across $\mathcal{R}$. More analyses on the effectiveness of the unequalness score can be found in the supplementary material. We plot in \cref{fig:realscore} the unequalness scores according to the number of real images on CIFAR-10~\cite{cifar}. We can see from this figure that the unequalness score of real images decreases rapidly, confirming the bilateral equivalence. Although the bilateral equivalence is an inherent characteristic of real datasets, we conjecture that it would prevent distilling fine-grained details into a small set of synthetic images. In particular, we have found that a synthetic dataset is highly likely to encode discriminative parts of objects on both the left and right halves of its images. This is because the discriminative parts could appear on both sides in a real dataset, and they provide strong supervisory signals at training time. Distilling discriminative parts only into the synthetic dataset however degrades performance, since fine-grained details provide more clues for recognizing subtle differences between objects.

\subsection{FYI}
\label{sec:fyi}
\begin{figure*}[t]
    \captionsetup{font={small}}
    \begin{center}
       \begin{subfigure}{\linewidth}
          \centering
            \captionsetup{justification=centering}
            \subcaptionbox{DSA~\cite{dsa}, 1 IPC}{
            \includegraphics[width=0.23\linewidth]{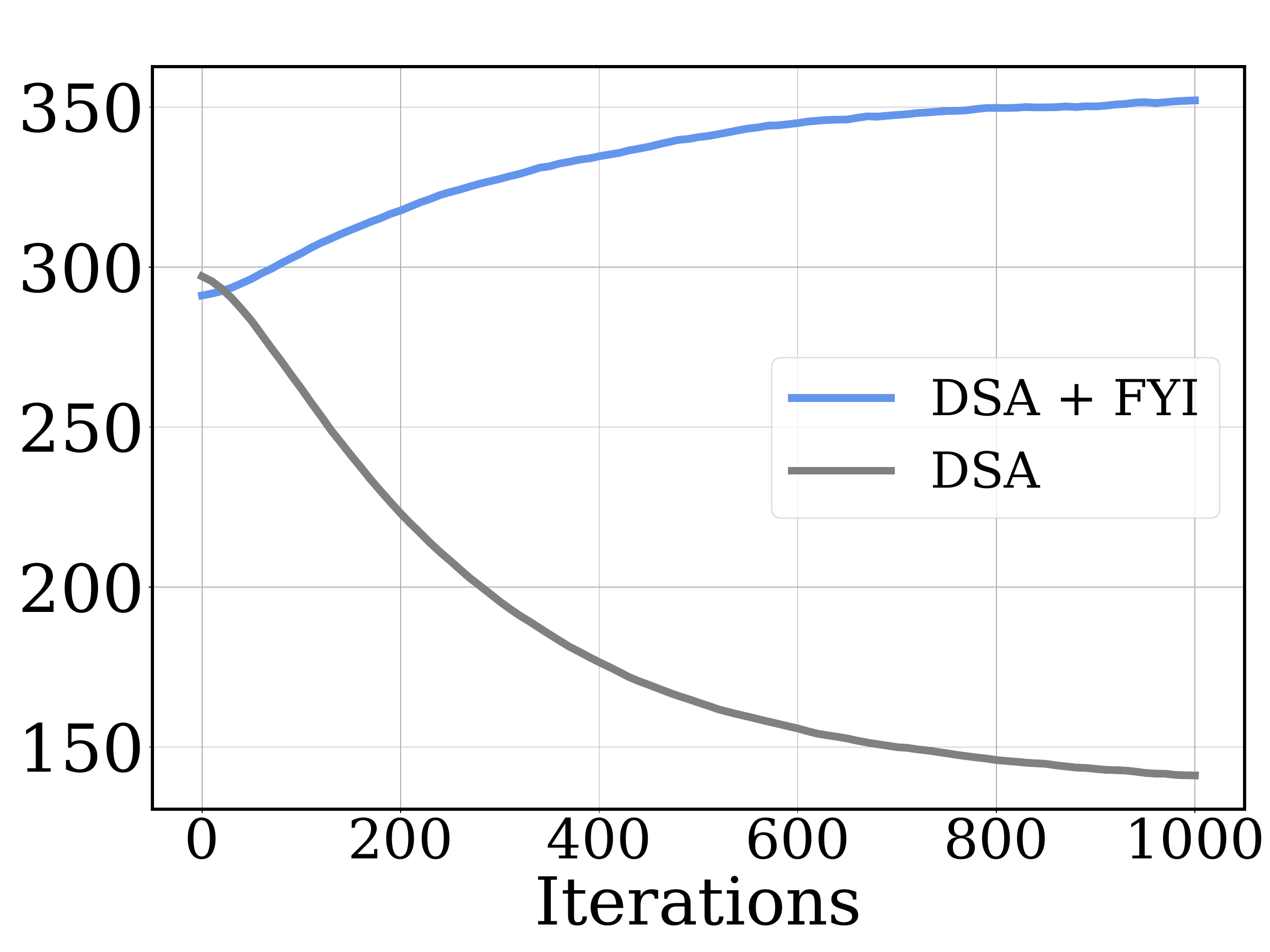}}
            \subcaptionbox{DSA, 10 IPC}{
            \includegraphics[width=0.23\linewidth]{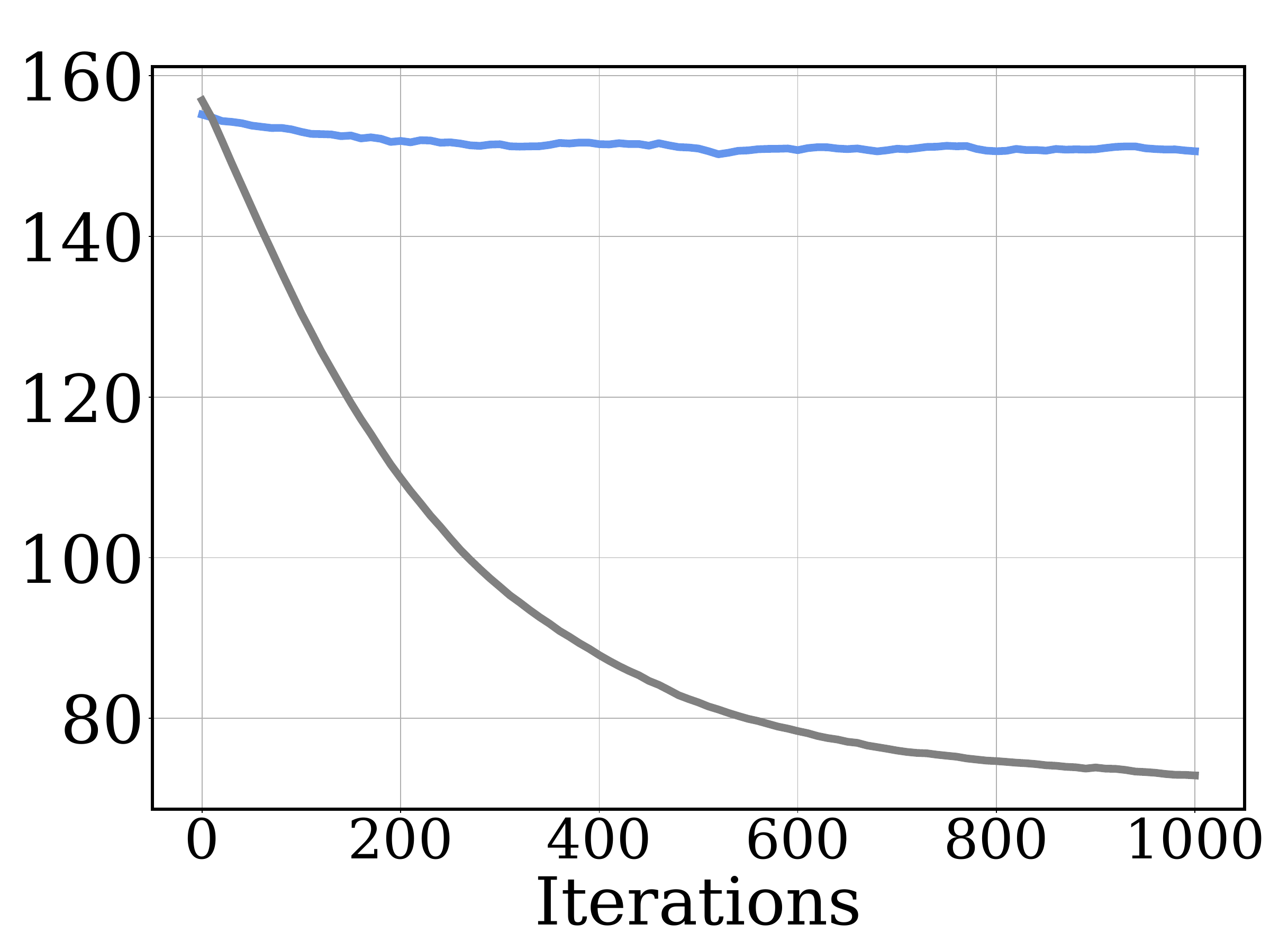}}
            \subcaptionbox{DM~\cite{dm}, 1 IPC}{
            \includegraphics[width=0.23\linewidth]{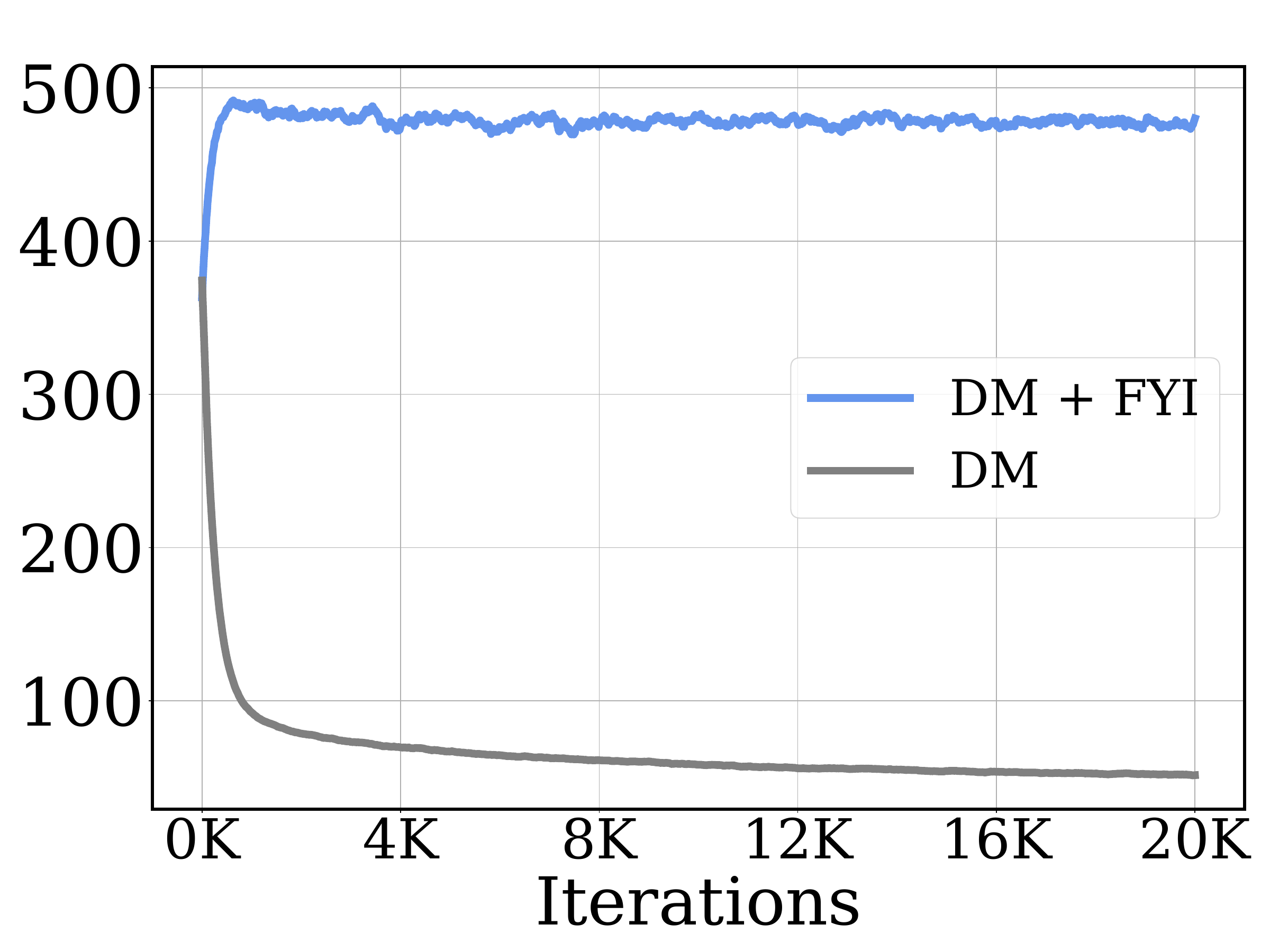}}
            \subcaptionbox{DM, 10 IPC}{
            \includegraphics[width=0.23\linewidth]{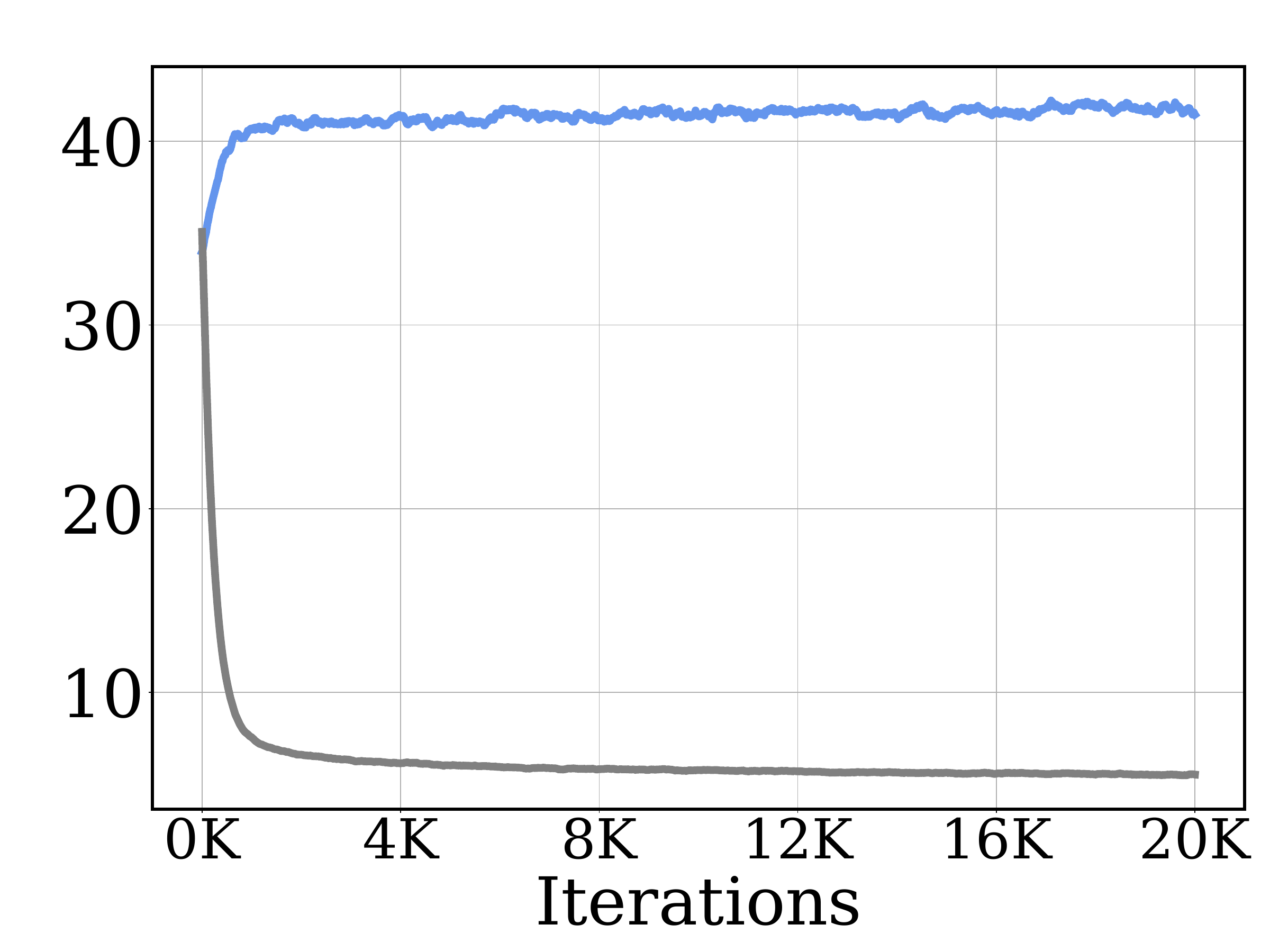}}
       \end{subfigure}
    \end{center}
    \vspace{-0.5cm}
       \caption{The bilateral equivalence of synthetic datasets with and without FYI on CIFAR-100~\cite{cifar}. We compute the unequalness score of synthetic images during training for (a-b) DSA~\cite{dsa} and (c-d) DM~\cite{dm}. FYI achieves a higher unequalness score compared to the vanilla methods during training, implying that it enables encoding different semantics on different halves of images. More experiments for different datasets, methods, and compression ratios can be found in the supplementary material.}
    \label{fig:verifications}
\end{figure*}
\begin{figure*}[t]
    \captionsetup{font={small}}
    \begin{center}
       \begin{subfigure}{\linewidth}
          \centering
            \captionsetup{justification=centering}
            \subcaptionbox{DSA~\cite{dsa}, 1 IPC}{
            \includegraphics[width=0.23\linewidth]{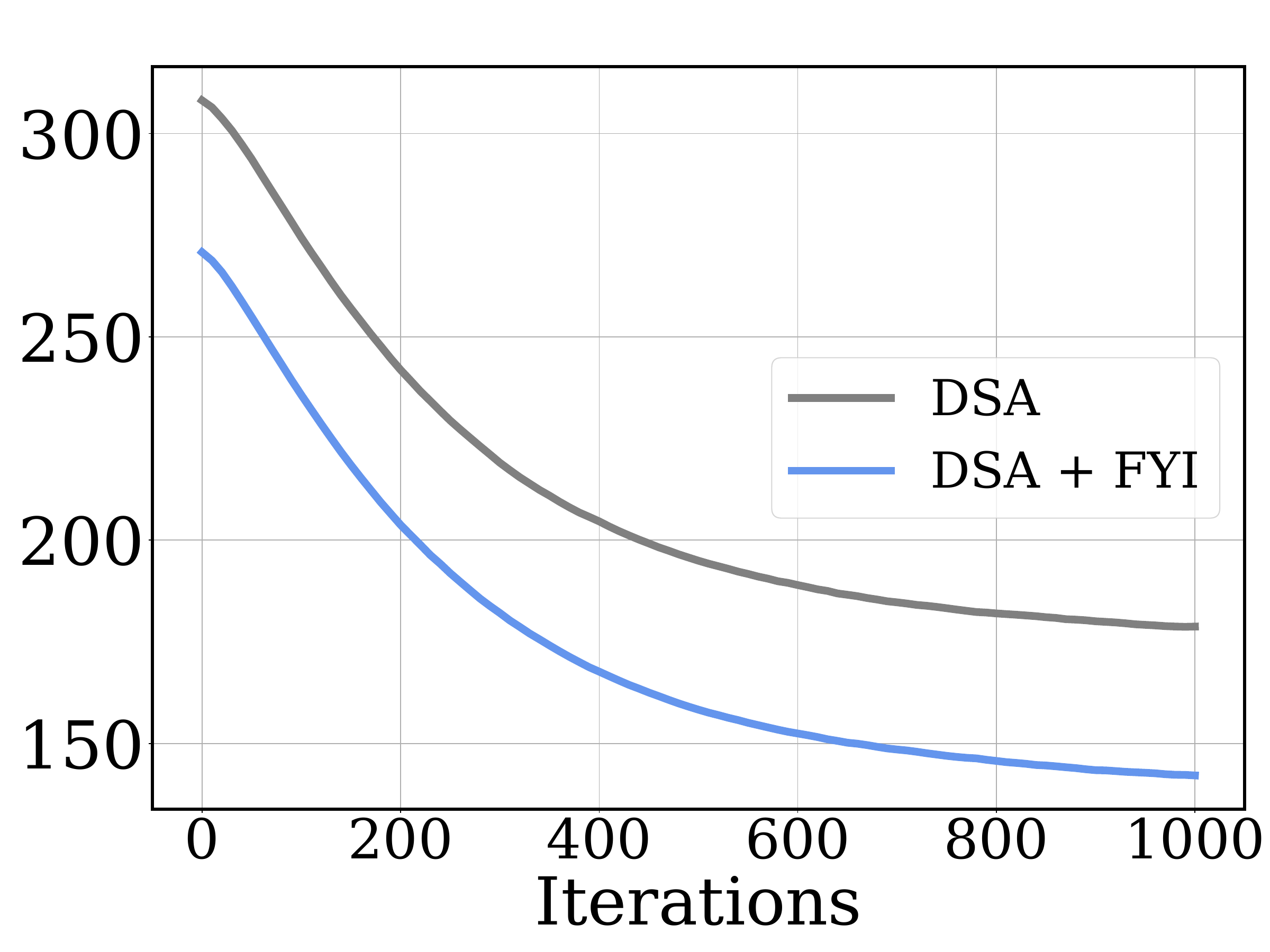}}
            \subcaptionbox{DSA, 10 IPC}{
            \includegraphics[width=0.23\linewidth]{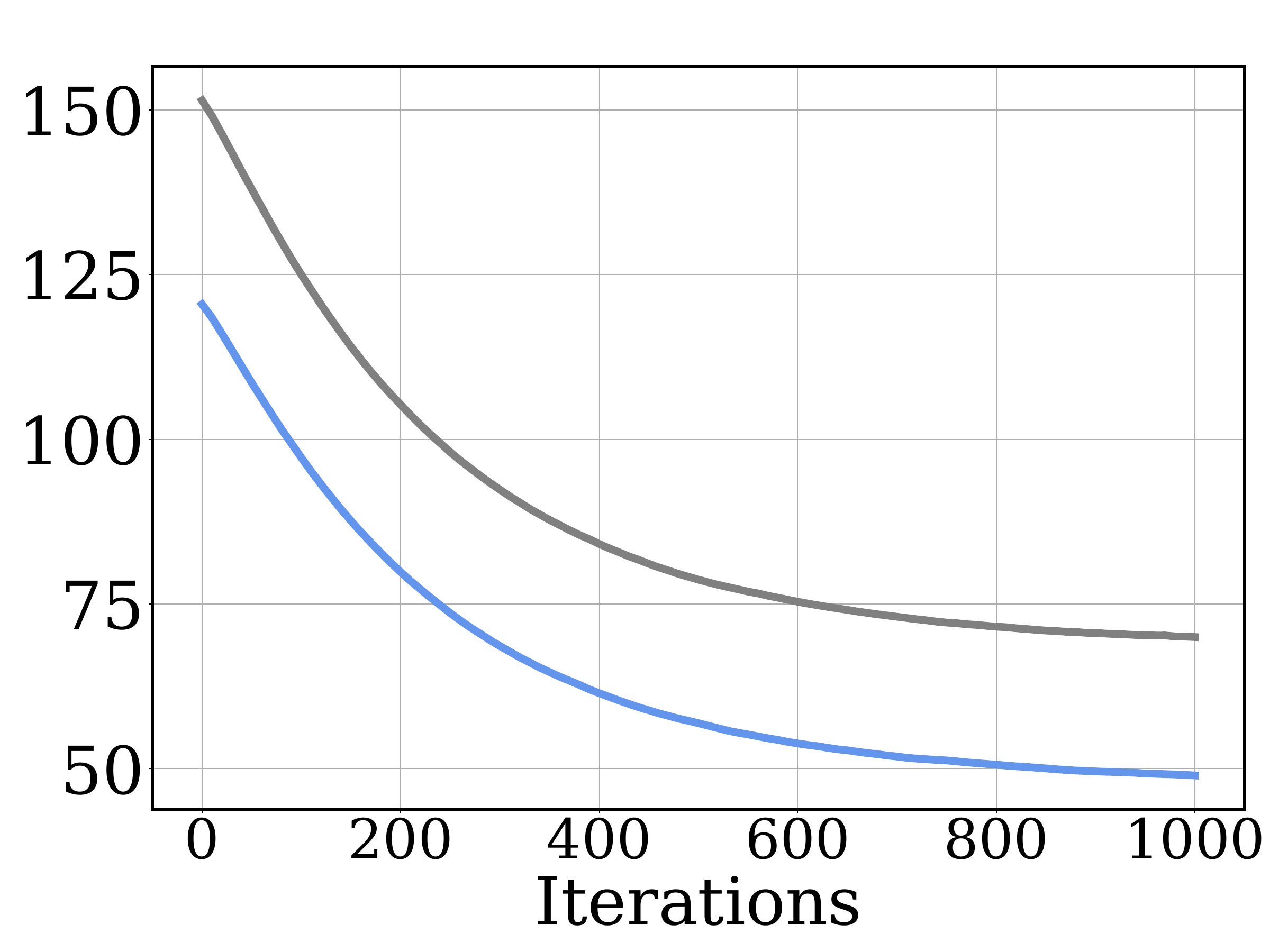}}
            \subcaptionbox{DM~\cite{dm}, 1 IPC}{
            \includegraphics[width=0.23\linewidth]{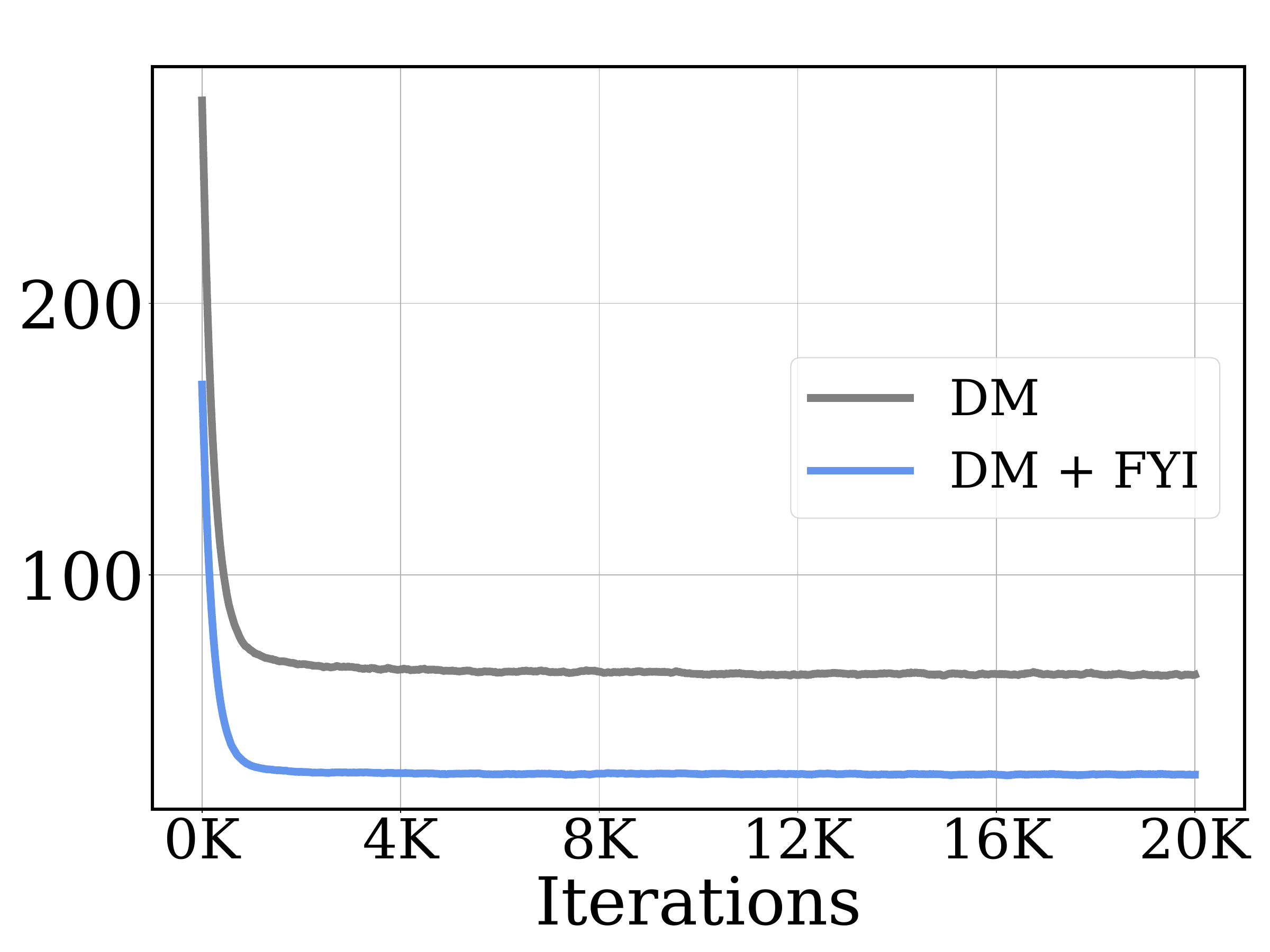}}
            \subcaptionbox{DM, 10 IPC}{
            \includegraphics[width=0.23\linewidth]{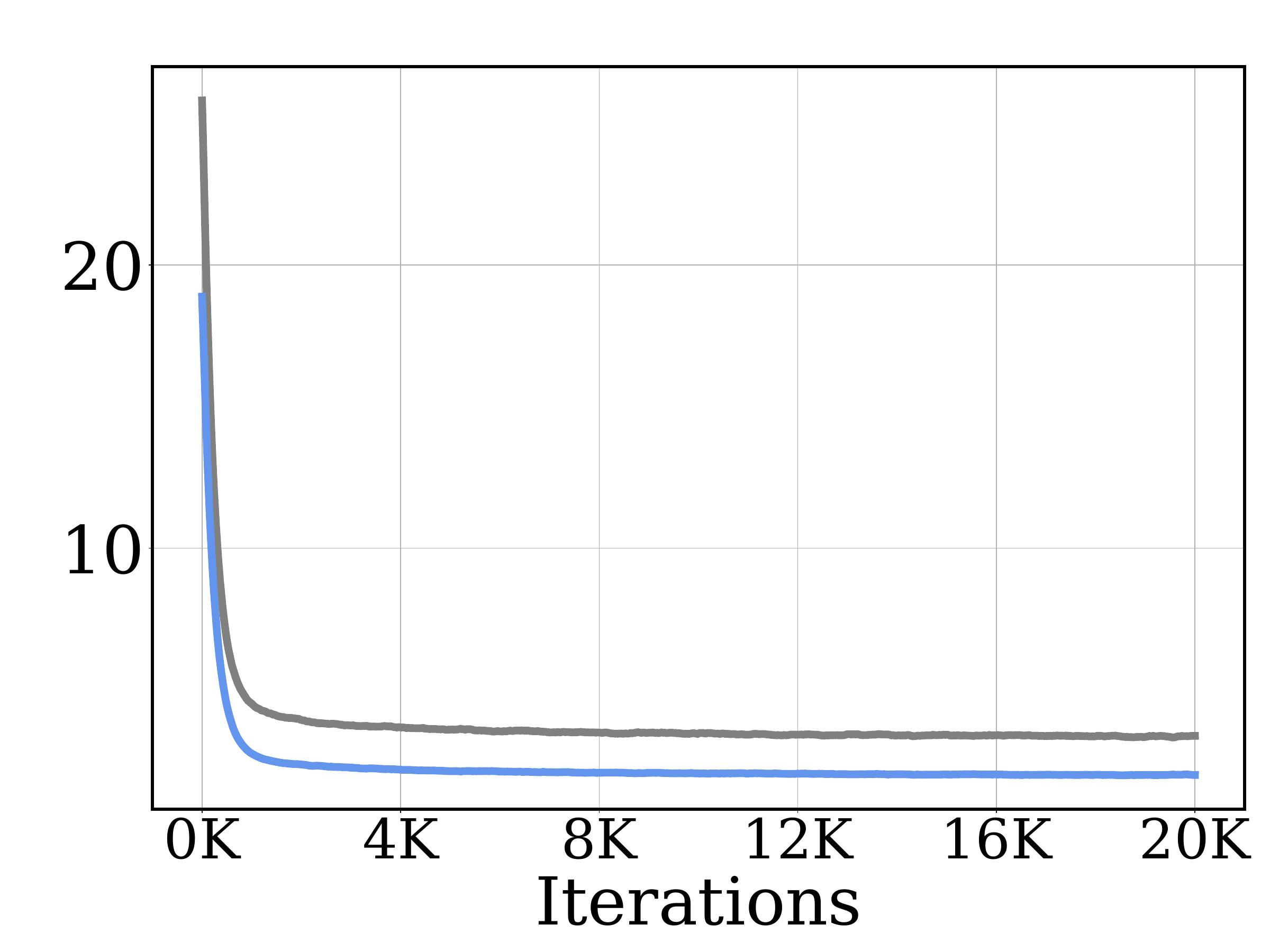}}
       \end{subfigure}
    \end{center}
    \vspace{-0.5cm}
       \caption{Plots of training losses in \cref{eq:1} or \cref{eq:6}, computed using (a-b) DSA~\cite{dsa} and (c-d) DM~\cite{dm}, on CIFAR-100~\cite{cifar}. FYI provides lower training losses compared to the vanilla methods consistently, which indicates that the distance between synthetic and real datasets is minimized more effectively by incorporating flipped counterparts of synthetic images into the dataset distillation process. More experiments can be found in the supplementary material.}
    \label{fig:loss}
\end{figure*}
We propose a surprisingly simple yet effective approach, dubbed FYI, that allows synthetic images to encode both discriminative parts and fine-grained details of objects (\cref{fig:overview}). To be specific, we propose to optimize synthetic images along with their flipped counterparts. Concretely, FYI first concatenates synthetic images with the flipped counterparts as follows:
\begin{equation}
    \small
    \label{eq:5}
    \mathcal{A}_{c} = \mathcal{S}_c \cup \text{Flip}(\mathcal{S}_c),
\end{equation}
where we denote by $\cup$ a batch-wise concatenation. It then exploits the augmented set of synthetic images to compute the following objective:
\begin{equation}
    \small
    \label{eq:6}
    \mathcal{L}_{\text{FYI}} = \mathbb{E}_{\theta\sim P_\theta}\Bigl[\sum_c D_\theta(\mathcal{T}_c, \mathcal{A}_{c})\Bigr].
\end{equation}
Since the flipping operation and the batch-wise concatenation are differentiable, we can update synthetic images as follows:
\begin{equation}
    \small
    \label{eq:7}
    \mathcal{S}_c \leftarrow \mathcal{S}_c - \eta \frac{\partial D_\theta(\mathcal{T}_c, \mathcal{A}_c)}{\partial \mathcal{A}_{c}} \frac{\partial \mathcal{A}_{c}}{\partial \mathcal{S}_c}.
\end{equation}
Note that FYI can be applied to any existing dataset distillation methods, since it does not modify network architectures and training objectives. We show in \cref{fig:verifications} that the unequalness scores of synthetic datasets with and without using FYI on CIFAR-100~\cite{cifar}. We can see that leveraging flipped images for the synthesis is very effective to avoid encoding duplicated patterns. Specifically, DSA~\cite{dsa} and DM~\cite{dm} without FYI show a rapid decrease of the unequalness score during training, indicating that both left and right halves of the synthetic images contain similar patterns (See \cref{fig:teaser}(b)). On the other hand, the score does not decrease for FYI, since it helps to capture discriminative parts of objects and fine-grained details (See \cref{fig:teaser}(c)). It is worth noting that the score gap is more significant if we synthesize a single image for dataset distillation~(\ie,~1 IPC). This suggests that FYI would be even more effective when a very limited number of synthetic images are affordable. To further demonstrate how FYI works, we show in \cref{fig:loss} the training loss of dataset distillation. We can see that FYI provides much lower training losses during image synthesis. This indicates that synthesized images using FYI better capture diverse semantics of real images, compared to the vanilla methods. Note that synthesizing an image with FYI is conditioned on both other synthetic images and flipped counterparts of all the images. In this context, FYI enables encoding different semantics on the left and right sides of synthetic images. We summarize in \cref{alg:algorithm} the overall dataset distillation process using FYI on top of DM.

\begin{algorithm}[t]
	\caption{Learning synthetic images using DM with FYI.}
	\label{alg:algorithm}
	\small
	\begin{algorithmic}[1]
      \Require Number of outer loop iterations $K$, number of classes $C$, parameter distribution $P_\theta$
      \INPUT Real dataset~$\mathcal{T}=\bigcup_c \mathcal{T}_c$.
      \State Initialize a synthetic dataset $\mathcal{S}=\bigcup_c \mathcal{S}_c$.
      \For{$k=0$ to $K-1$}
         \State Sample network parameters $\theta$ from $P_\theta$.
         \For{$c=0$ to $C-1$}
            \State FYI: flip-and-concatenate the synthetic images using Eq.~\eqref{eq:5}.
            \State Calculate $D_\theta(\mathcal{T}_c, \mathcal{A}_c)$
            \State Update $\mathcal{S}_c$ using Eq.~\eqref{eq:7}.
         \EndFor
      \EndFor
   \end{algorithmic}
\end{algorithm}

\begin{table*}[t]
  \caption{Quantitative comparison on the test set of CIFAR-10/100~\cite{cifar} and the validation split of Tiny-ImageNet~\cite{tiny}. We report the average top-1 accuracy (\%) with standard deviations.}
  \resizebox{\textwidth}{!}{%
  \begin{tabular}{c|ccc|ccc|ccc}
  \hline
  Dataset
   &
    \multicolumn{3}{c|}{CIFAR 10} &
    \multicolumn{3}{c|}{CIFAR 100} &
    \multicolumn{3}{c}{Tiny-ImageNet} \\\hline
  IPC &
    1 &
    10 &
    50 &
    1 &
    10 &
    50 &
    1 &
    10 &
    50 \\
  Ratio (\%) &
    0.02 &
    0.2 &
    1 &
    0.2 &
    2 &
    10 &
    0.2 &
    2 &
    10 \\ \hline \hline
    CAFE~\cite{cafe} &
      30.3 (1.1) &
      46.3 (0.6) &
      55.5 (0.6) &
      12.9 (0.3) &
      27.8 (0.3) &
      37.9 (0.3) &
      - &
      - &
      - \\
    CAFE+DSA &
      31.6 (0.8) &
      50.9 (0.5) &
      62.3 (0.4) &
      14.0 (0.3) &
      31.5 (0.2) &
      42.9 (0.2) &
      - &
      - &
      - \\
    DataDAM~\cite{datadam} &
      32.0 (1.2) &
      54.2 (0.8) &
      67.0 (0.4) &
      14.5 (0.5) &
      34.8 (0.5) &
      49.4 (0.3) &
      8.3 (0.4) &
      18.7 (0.3) &
      28.7 (0.3) \\ 
        DC~\cite{dc} &
        28.3 (0.5) &
        44.9 (0.5) &
        53.9 (0.5) &
        12.8 (0.3) &
        25.2 (0.3) &
        - &
        - &
        - &
        - \\
        \rowcolor[HTML]{DAE8FC} 
      DC + FYI&
      30.0 (0.5) &
          49.5 (0.5) &
          54.6 (0.6) &
          14.4 (0.3) &
          29.2 (0.3) &
          - &
          - &
          - &
          - \\\hline
      DSA~\cite{dsa} &
        28.8 (0.7) &
        52.1 (0.5) &
        60.6 (0.5) &
        13.9 (0.3) &
        32.3 (0.3) &
        42.8 (0.4) &
        - &
        - &
        - \\
        \rowcolor[HTML]{DAE8FC} 
      DSA + FYI &
      30.6 (0.7) &
      54.7 (0.5) &
      63.7 (0.5) &
      16.0 (0.3) &
      35.0 (0.3) &
      45.4 (0.4) &
          - &
          - &
          - \\\hline
          IDC~\cite{idc} &
              50.1 (0.4) &
              67.5 (0.5) &
              74.5 (0.1) &
              28.1 (0.2) &
              45.0 (0.3) &
              - &
              - &
              - &
              - \\
              \rowcolor[HTML]{DAE8FC} 
            IDC + FYI &
            52.5 (0.4) &
            68.1 (0.3) &
            75.2 (0.0) &
            28.9 (0.1) &
            45.8 (0.2) &
                - &
                - &
                - &
                - \\\hline
                DM~\cite{dm} &
                  26.0 (0.8) &
                  48.9 (0.6) &
                  63.0 (0.4) &
                  11.4 (0.3) &
                  29.7 (0.3) &
                  43.6 (0.4) &
                  3.9 (0.2) &
                  12.9 (0.4) &
                  24.1 (0.3) \\
                  \rowcolor[HTML]{DAE8FC} 
                DM + FYI &
                28.7 (1.2) &
                53.1 (0.6) &
                64.2 (0.4) &
                13.3 (0.3) &
                32.3 (0.4) &
                45.6 (0.4) &
                4.6 (0.2) &
                16.6 (0.3) &
                25.6 (0.4) \\\hline
      MTT~\cite{mtt} &
        46.3 (0.8) &
        65.3 (0.7) &
        71.6 (0.2) &
        24.3 (0.3) &
        40.1 (0.4) &
        47.7 (0.2) &
        9.0 (0.3)  &
        24.5 (0.4) &
        29.7 (0.4) \\
        \rowcolor[HTML]{DAE8FC} 
      MTT + FYI &
      47.2 (0.4) &
      68.2 (0.3) &
      74.0 (0.3) &
      28.2 (0.3) &
      41.6 (0.2) &
      48.2 (0.2) &
      10.4 (0.3)  &
      25.2 (0.5) &
      30.1 (0.3) \\
      FTD~\cite{ftd} &
        46.8 (0.3) &
        66.6 (0.3) &
        73.8 (0.2) &
        25.2 (0.2) &
        43.4 (0.3) &
        50.7 (0.3) &
        10.4 (0.3)  &
        26.4 (0.1) &
        - \\
        \rowcolor[HTML]{DAE8FC} 
      FTD + FYI &
      49.5 (1.0) &
      68.1 (0.5) &
      74.5 (0.3) &
      28.7 (0.7) &
      44.7 (0.3) &
      50.8 (0.4) &
      11.6 (0.1)  &
      26.8 (0.3) &
      - \\\hline
  Full dataset &
    \multicolumn{3}{c|}{84.8 (0.1)} &
    \multicolumn{3}{c|}{56.2 (0.3)} &
    \multicolumn{3}{c}{37.6 (0.4)} \\ \hline
  \end{tabular}%
  }
  \label{tab:sota}
  \end{table*}
\begin{table*}[t]
  \caption{Quantitative comparison on the validation split of ImageNet subsets~\cite{insub} for 1 and 10 IPC settings. The numbers in the brackets indicate the standard deviations.}
    \resizebox{\textwidth}{!}{%
    \begin{tabular}{c|cc|cc|cc|cc|cc|cc}
    \hline
    Dataset &
      \multicolumn{2}{c|}{ImageNette} &
      \multicolumn{2}{c|}{ImageWoof} &
      \multicolumn{2}{c|}{ImageFruit} &
      \multicolumn{2}{c|}{ImageMeow} &
      \multicolumn{2}{c|}{ImageSquawk} &
      \multicolumn{2}{c}{ImageYellow} \\
    IPC &
      1 &
      10 &
      1 &
      10 &
      1 &
      10 &
      1 &
      10 &
      1 &
      10 &
      1 &
      10 \\ \hline \hline
      FTD~\cite{ftd} &
        52.2 (1.0) &
        67.7 (0.7) &
        30.1 (1.0) &
        38.8 (1.4) &
        29.1 (0.9) &
        44.9 (1.5) &
        33.8 (1.5) &
        43.3 (0.6) &
        - &
        - &
        - &
        - \\
    MTT~\cite{mtt} &
      47.7 (0.9) &
      63.0 (1.3) &
      28.6 (0.8) &
      35.8 (1.8) &
      26.6 (0.8) &
      40.3 (1.3) &
      30.7 (1.6) &
      40.4 (2.2) &
      39.4 (1.5) &
      52.3 (1.0) &
      45.2 (0.8) &
      60.0 (1.5) \\
      \rowcolor[HTML]{DAE8FC}
    MTT + FYI &
       52.4 (2.6) &
       68.4 (1.6) &
       30.6 (1.1) &
       40.3 (0.7) &
       30.1 (1.4) &
       46.0 (1.1) &
       33.5 (1.6) &
       46.8 (0.8) &
       42.6 (0.6) &
       61.6 (1.6) &
       48.4 (1.0) &
       66.4 (1.9) \\ \hline
    Full dataset &
      \multicolumn{2}{c|}{87.4 (1.0)} &
      \multicolumn{2}{c|}{67.0 (1.3)} &
      \multicolumn{2}{c|}{63.9 (2.0)} &
      \multicolumn{2}{c|}{66.7 (1.1)} &
      \multicolumn{2}{c|}{87.5 (0.3)} &
      \multicolumn{2}{c}{84.4 (0.6)} \\ \hline
    \end{tabular}%
    }
    \label{tab:subset}
    \end{table*}
\section{Experiments}
In this section, we describe our implementation details~(\cref{sec:exp_implementation}) and provide quantitative and qualitative comparisons of our approach with state-of-the-art methods~(\cref{sec:exp_results}). We also present extensive analyses on our approach~(\cref{sec:exp_discussion}). Please refer to the supplementary material for more results including applications to continual learning and NAS.

\subsection{Implementation details}
\label{sec:exp_implementation}
\subsubsection{Datasets.} We perform experiments on standard benchmarks: CIFAR-10/100~\cite{cifar}, Tiny-ImageNet~\cite{tiny}, and ImageNet~\cite{imagenet}. The CIFAR-10/100 datasets consist of 50K training and 10K test images of size 32$\times$32 for 10 and 100 object classes, respectively. The Tiny-ImageNet dataset provides 100K training and 10K validation images of size 64$\times$64 for 200 classes. Following~\cite{mtt}, we use six subsets of ImageNet, where all images are resized to the size of 128$\times$128. Each subset contains approximately 12K training and 500 validation images for 10 classes. For evaluation, we use the validation splits for Tiny-ImageNet and ImageNet, following the experimental protocol in~\cite{mtt}.

\subsubsection{Training and evaluation.} We apply FYI to several state-of-the-art methods: DC~\cite{dc}, DSA~\cite{dsa}, IDC~\cite{idc}, DM~\cite{dm}, MTT~\cite{mtt}, and FTD~\cite{ftd}. We follow the training details of each method. To be specific, we use a ConvNet~\cite{conv} architecture for both distillation and retraining processes. ConvNet consists of 3, 4, and 5 blocks on CIFAR-10/100~\cite{cifar}, Tiny-ImageNet~\cite{tiny}, and ImageNet~\cite{imagenet}, respectively, where each block contains a 3$\times$3 convolutional layer with 128 channels followed by a ReLU~\cite{alexnet} activation and a 2$\times$2 average pooling layer. We halve the batch size of synthetic images for MTT and FTD before applying FYI in order to maintain computational costs of original methods. For evaluation, we retrain ConvNet with the synthesized images for 1K epochs using the SGD optimizer with a learning rate of 0.01, a momentum of 0.9, and a weight decay of 5$e$-4. The learning rate is adjusted by the step schedule. We use 6 operations for data augmentation, namely, crop, color jitters~\cite{alexnet}, cutout~\cite{cutout}, flip, scale, and rotate. For IDC, we also apply CutMix~\cite{cutmix} following the original work. Note that we do not concatenate flipped images as in Eq.~\eqref{eq:5} during retraining for a fair comparison. We measure the classification accuracy on the test or validation splits of each dataset, and report average accuracies using 100, 100, 3, 25, and 5 different random seeds for DC, DSA, IDC, DM, and MTT, respectively. We provide more details in the supplementary material.

\begin{figure*}[t]
    \captionsetup{font={small}}
    \begin{center}
 
       \includegraphics[width=0.12\linewidth]{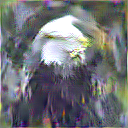}\hspace{0.005\linewidth}%
       \includegraphics[width=0.12\linewidth]{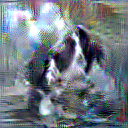}\hspace{0.005\linewidth}%
       \includegraphics[width=0.12\linewidth]{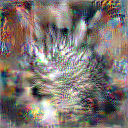}\hspace{0.005\linewidth}%
       \includegraphics[width=0.12\linewidth]{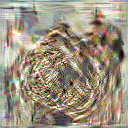}\hspace{0.005\linewidth}%
       \includegraphics[width=0.12\linewidth]{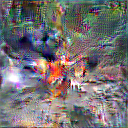}\hspace{0.005\linewidth}%
       \includegraphics[width=0.12\linewidth]{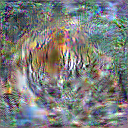}\hspace{0.005\linewidth}%
       \includegraphics[width=0.12\linewidth]{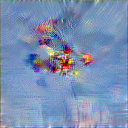}\hspace{0.005\linewidth}%
       \includegraphics[width=0.12\linewidth]{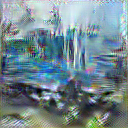}
 
       \vspace{0.003\linewidth}
 
       \includegraphics[width=0.12\linewidth]{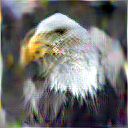}\hspace{0.005\linewidth}%
       \includegraphics[width=0.12\linewidth]{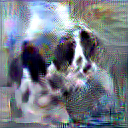}\hspace{0.005\linewidth}%
       \includegraphics[width=0.12\linewidth]{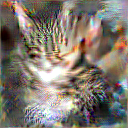}\hspace{0.005\linewidth}%
       \includegraphics[width=0.12\linewidth]{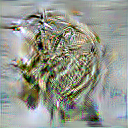}\hspace{0.005\linewidth}%
       \includegraphics[width=0.12\linewidth]{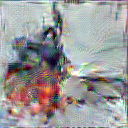}\hspace{0.005\linewidth}%
       \includegraphics[width=0.12\linewidth]{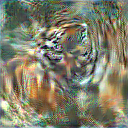}\hspace{0.005\linewidth}%
       \includegraphics[width=0.12\linewidth]{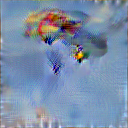}\hspace{0.005\linewidth}%
       \includegraphics[width=0.12\linewidth]{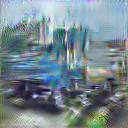}
 
    \end{center}
    \vspace{-1em}
    \caption{Qualitative comparison between MTT~\cite{mtt} (top) and MTT+FYI (bottom) on ImageNet~\cite{imagenet}: Bald eagle, English springer, tabby cat, French horn, chainsaw, tiger, parachute, and garbage truck classes. We observe that FYI improves MTT to encode fine-grained details of objects.}
    \label{fig:imagenet}
 \end{figure*}
 
 \begin{figure*}[t]
    \captionsetup{font={small}}
    \begin{center}
       \includegraphics[width=0.097\linewidth]{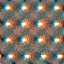}\hspace{0.002\linewidth}%
       \includegraphics[width=0.097\linewidth]{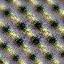}\hspace{0.002\linewidth}%
       \includegraphics[width=0.097\linewidth]{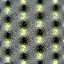}\hspace{0.002\linewidth}%
       \includegraphics[width=0.097\linewidth]{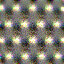}\hspace{0.002\linewidth}%
       \includegraphics[width=0.097\linewidth]{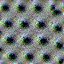}\hspace{0.002\linewidth}%
       \includegraphics[width=0.097\linewidth]{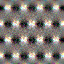}\hspace{0.002\linewidth}%
       \includegraphics[width=0.097\linewidth]{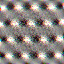}\hspace{0.002\linewidth}%
       \includegraphics[width=0.097\linewidth]{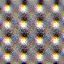}\hspace{0.002\linewidth}%
       \includegraphics[width=0.097\linewidth]{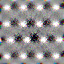}\hspace{0.002\linewidth}%
       \includegraphics[width=0.097\linewidth]{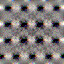}
       \vspace{0.01\linewidth}
       \includegraphics[width=0.097\linewidth]{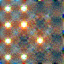}\hspace{0.002\linewidth}%
       \includegraphics[width=0.097\linewidth]{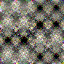}\hspace{0.002\linewidth}%
       \includegraphics[width=0.097\linewidth]{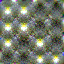}\hspace{0.002\linewidth}%
       \includegraphics[width=0.097\linewidth]{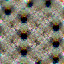}\hspace{0.002\linewidth}%
       \includegraphics[width=0.097\linewidth]{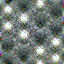}\hspace{0.002\linewidth}%
       \includegraphics[width=0.097\linewidth]{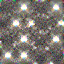}\hspace{0.002\linewidth}%
       \includegraphics[width=0.097\linewidth]{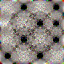}\hspace{0.002\linewidth}%
       \includegraphics[width=0.097\linewidth]{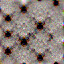}\hspace{0.002\linewidth}%
       \includegraphics[width=0.097\linewidth]{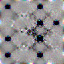}\hspace{0.002\linewidth}%
       \includegraphics[width=0.097\linewidth]{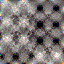}
    \end{center}
    \vspace{-1em}
    \caption{Qualitative comparison of DM~\cite{dm} (top) and DM+FYI (bottom) on the first 10 classes of Tiny-ImageNet~\cite{tiny}. FYI synthesizes images in various patterns, whereas the vanilla method duplicates patterns in the left and right halves of images.}
    \label{fig:tiny}
    \vspace{-.5cm}
 \end{figure*}
\begin{figure}[t]
    \begin{minipage}{0.46\linewidth}
        \captionsetup{font={small}}
        \begin{center}
            \begin{subfigure}{\linewidth}
               \centering
               \includegraphics[width=0.18\linewidth]{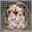}\hspace{0.002\linewidth}%
               \includegraphics[width=0.18\linewidth]{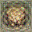}\hspace{0.002\linewidth}%
               \includegraphics[width=0.18\linewidth]{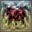}\hspace{0.002\linewidth}%
               \includegraphics[width=0.18\linewidth]{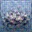}\hspace{0.002\linewidth}%
               \includegraphics[width=0.18\linewidth]{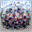}
               \vspace{0.01\linewidth}
               \includegraphics[width=0.18\linewidth]{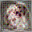}\hspace{0.002\linewidth}%
               \includegraphics[width=0.18\linewidth]{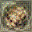}\hspace{0.002\linewidth}%
               \includegraphics[width=0.18\linewidth]{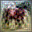}\hspace{0.002\linewidth}%
               \includegraphics[width=0.18\linewidth]{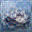}\hspace{0.002\linewidth}%
               \includegraphics[width=0.18\linewidth]{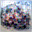}
            \end{subfigure}
        \end{center}
        \caption{Qualitative comparison of synthetic images trained on CIFAR-10~\cite{cifar}. We visualize synthetic images from the following object categories: dog, frog, horse, ship, and truck. Top: Synthesized images using DSA contain discriminative parts repeatedly (\eg, heads of horses). Bottom: Applying FYI to DSA helps to capture different parts of objects (\eg, the tail of a horse).}
        \label{fig:cifar10}
    \end{minipage}\hfill
    \begin{minipage}{0.5\linewidth}
        \captionsetup{font={small}}
        \captionof{table}{Comparison of the top-1 accuracy (\%) for different network architectures. We synthesize images using ConvNet~\cite{conv} on CIFAR-10~\cite{cifar} with 50 IPC, and use them to train ConvNet~\cite{conv}, VGG-11~\cite{vgg}, and ResNet-18~\cite{resnet}. We use DC~\cite{dc}, DSA~\cite{dsa}, DM~\cite{dm}, and MTT~\cite{mtt} for image synthesis. We report the standard deviations in the brackets.}
        \resizebox{\columnwidth}{!}{%
        \begin{tabular}{cccc}
        \hline
        \multicolumn{1}{l}{} & ConvNet~\cite{conv}               & VGG-11~\cite{vgg}              & ResNet-18~\cite{resnet}           \\ \hline
        DC~\cite{dc}                   & 53.9 (0.5)            & 38.8 (1.1)          & 20.9 (1.0)          \\
        \rowcolor[HTML]{DAE8FC} 
        DC+FYI               &  54.6 (0.6) &  40.8 (0.7) &  25.8 (0.8) \\
        DSA~\cite{dsa}                  & 60.6 (0.5)            & 51.4 (1.0)          & 47.8 (0.9)          \\
        \rowcolor[HTML]{DAE8FC} 
        DSA+FYI              &  63.7 (0.5) &  56.2 (0.6) &  52.9 (0.8) \\
        DM~\cite{dm}                   & 63.0 (0.4)            & 57.4 (0.8)          & 52.9 (0.4)          \\
        \rowcolor[HTML]{DAE8FC} 
        DM+FYI               &  64.2 (0.4) &  59.6 (0.5) &  56.1 (0.8) \\
        MTT~\cite{mtt}                  & 71.6 (0.2)            & 61.5 (0.5)          & 58.7 (0.2)          \\
        \rowcolor[HTML]{DAE8FC} 
        MTT+FYI              &  74.0 (0.3) &  65.7 (0.5) &  61.1 (0.6) \\ \hline
        \end{tabular}%
        }
        \label{tab:cross}
    \end{minipage}
    \vspace{-.5cm}
  \end{figure}

\subsection{Results}
\label{sec:exp_results}
\subsubsection{Quantitative results.} 
We compare in Table~\ref{tab:sota} results of state-of-the-art methods on CIFAR-10/100~\cite{cifar} and Tiny-ImageNet~\cite{tiny} with varying numbers of synthetic images. From this table, we have three findings: (1) FYI gives remarkable gains over DC~\cite{dc}, DM~\cite{dm}, and MTT~\cite{mtt} consistently. This demonstrates that FYI can be easily applied to different types of training objectives (\ie, distribution~\cite{dm}, gradient~\cite{dc}, and trajectory matching~\cite{mtt}) to improve the distillation performance. (2) All methods using FYI provide better results, especially in challenging scenarios (\emph{e.g.}, 1 IPC). This suggests that the problem caused by the bilateral equivalence becomes severe, as the number of synthetic images becomes smaller. Our FYI mitigates the problem effectively, achieving the accuracy gains of 2.7\%, 3.5\%, and 1.2\% over FTD~\cite{ftd} for the 1 IPC case on CIFAR-10, CIFAR-100 and Tiny-ImageNet, respectively. (3) FYI brings large improvements over DSA~\cite{dsa} and IDC~\cite{idc}. This shows that FYI improves the performance of dataset distillation in a complementary manner to existing methods using data augmentation techniques. DSA applies the same data augmentation technique (\eg, rotate 10 degrees) to real and synthetic images before feeding them into networks. IDC enlarges the number of synthetic images by resizing low-resolution images, keeping the total storage budget. For example, IDC synthesizes 40 images for the 10 IPC setting but stores the same number of pixels as 10 real images. While FYI also exploits a data augmentation technique (\ie, horizontal flipping), it mitigates a different problem caused by the bilateral equivalence. Note that both DSA and IDC suffer from this problem, and FYI further improves the performance consistently.

\subsubsection{Qualitative results.} We show in \cref{fig:imagenet,fig:tiny,fig:cifar10} qualitative results obtained without (top) and with (bottom) FYI. Compared to the original methods~\cite{mtt,dm,dsa}, we can see that FYI provides synthetic images containing rich semantics, including discriminative parts of objects and fine-grained details. For example, \cref{fig:imagenet} shows that MTT~\cite{mtt} using FYI produces synthetic images containing fine-grained details such as the beak of a bald eagle (the first column) and the blade of a chainsaw (the fifth column). We can see from \cref{fig:tiny} that DM~\cite{dm} without FYI produces duplicated shapes (top), while using FYI avoids duplicating patterns on both the left and right sides of images (bottom). We can also observe in \cref{fig:cifar10} that FYI can also be effective in distilling low-resolution images.

\subsection{Discussion}
\label{sec:exp_discussion}

\subsubsection{Fine-grained classification.} We provide in Table~\ref{tab:subset} results of our method on subsets of ImageNet~\cite{imagenet} for two IPC cases. We can see that MTT~\cite{mtt} using FYI outperforms state-of-the-art methods~\cite{mtt,ftd} significantly on all subsets for all IPC settings, validating once again the effectiveness of the proposed FYI. In particular, the accuracy gains from FYI are 3.2\% and 9.3\% for 1 and 10 IPC cases on ImageSquawk. FYI removes duplicated patterns, while capturing fine-grained details~(See Fig.~\ref{fig:imagenet}), which is crucial for recognizing such an object.

\begin{table}[t]
    \caption{Quantitative comparison of the top-1 accuracy (\%) of FYI and its variants using different data augmentation techniques. We synthesize images on CIFAR-10~\cite{cifar} with 50 IPC. We report the standard deviations in the brackets.}
    \centering
    \resizebox{.7\textwidth}{!}{%
    \begin{tabular}{c|>{\centering\arraybackslash}m{2cm}>{\centering\arraybackslash}m{2cm}>{\centering\arraybackslash}m{ 2cm}>{\centering\arraybackslash}m{ 2cm}>{\centering\arraybackslash}m{ 2cm}}
    \hline
        & w/o augmentation  & Horizontal Flip~(FYI)        & Rotate                             & Scale                              & Vertical Flip                      \\ \hline
    DC~\cite{dc}  & 53.9 (0.5) & \textbf{54.6 (0.6)} & 52.5 (0.6)          & 36.1 (0.6)          & 32.2 (0.9) \\
    DSA~\cite{dsa} & 60.6 (0.5) & \textbf{63.7 (0.5)} & 62.1 (0.4)          & 55.9 (0.6)          & 46.5 (0.5)          \\
    DM~\cite{dm}  & 63.0 (0.4) & \textbf{64.2 (0.4)} & 63.3 (0.3) & 58.1 (0.5) & 57.6 (0.4) \\ \hline
    \end{tabular}%
    }
    \label{tab:ablation}
    \end{table}

\begin{figure}[t]
    \captionsetup{font={small}}
    \begin{center}
       \begin{subfigure}{\linewidth}
          \centering
          \captionsetup{justification=centering}
          \hspace{0.5cm}{
           \hspace{-0.5cm}\includegraphics[width=0.35\linewidth]{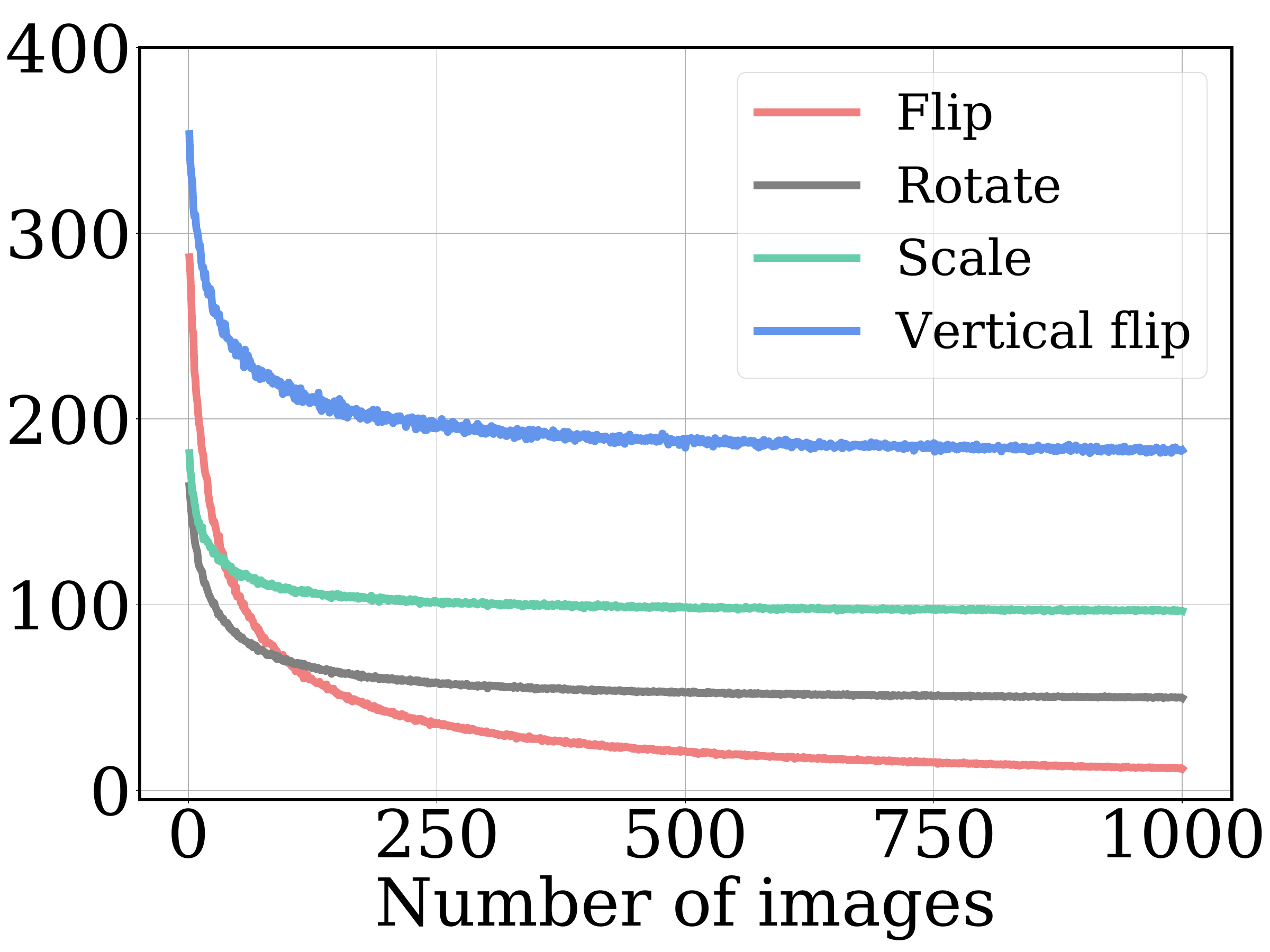}}
            \hspace{0.5cm}{
            \hspace{-0.5cm}\includegraphics[width=0.35\linewidth]{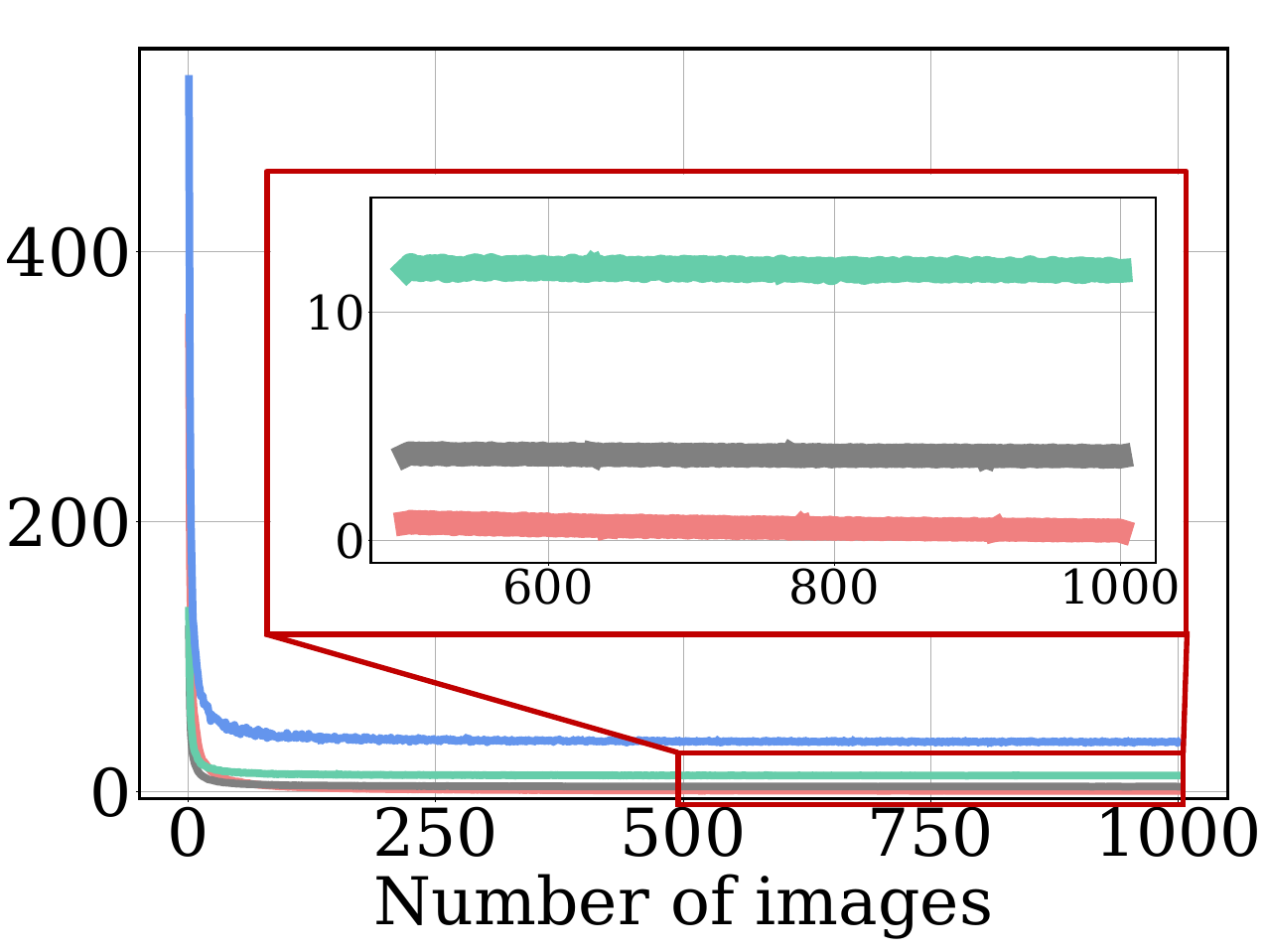}}
       \end{subfigure}
    \end{center}
    \vspace{-.5cm}
    \caption{The unequalness score and its variants on CIFAR-10~\cite{cifar}. For the variants, we replace \text{Flip} in~\cref{eq:4} with different data augmentation techniques. We use  (left) DC~\cite{dc} and (right) DM~\cite{dm} as a distance metric $D_\theta$. We  report the scores averaged over object classes, similar to the results in~\cref{fig:realscore}.}
    \label{fig:score_variants}
    \vspace{-.5cm}
\end{figure}
\subsubsection{Cross-architecture generalization.} We report in Table~\ref{tab:cross} the top-1 accuracy of network architectures that are unseen during the image synthesis. Specifically, we train synthetic images with ConvNet~\cite{conv} and use them to train VGG-11~\cite{vgg} and ResNet-18~\cite{resnet} for evaluation. We can see that FYI again provides remarkable improvements over the original methods consistently. This indicates that FYI helps to synthesize images encoding rich semantics robust to various network architectures effectively.

\begin{figure}[t]
    \captionsetup{font={small}}
    \begin{center}
       \begin{subfigure}{\linewidth}
          \centering
            \captionsetup{justification=centering}
            \subcaptionbox{DSA~\cite{dsa}, 1 IPC}{
            \includegraphics[width=0.48\linewidth]{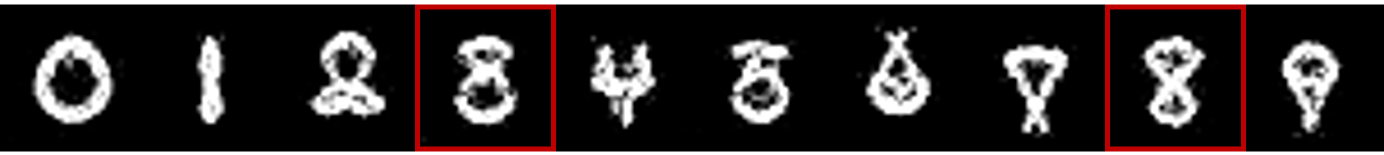}}
            \subcaptionbox{DSA + FYI, 1 IPC}{
            \includegraphics[width=0.48\linewidth]{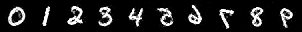}}
            \subcaptionbox{DSA, 2 IPC}{
            \includegraphics[width=0.48\linewidth]{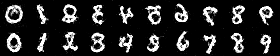}}
            \subcaptionbox{DSA + FYI, 2 IPC}{
            \includegraphics[width=0.48\linewidth]{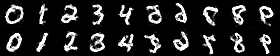}}
       \end{subfigure}
       \vspace{-1em}
    \end{center}
       \caption{Qualitative comparison of synthetic images trained on the extended MNIST~\cite{mnist} dataset. (a) Images synthesized using DSA~\cite{dsa} for 1 IPC. We observe all synthesized images are symmetric. (b) Applying FYI to DSA provides asymmetric images, with identifiable digits. (c) The vanilla DSA with 2 IPC still encodes similar semantics in the left and right halves of images. (d) DSA using FYI with 2 IPC shows that two synthetic images for the same digit capture different semantics effectively, while being asymmetric.}
    \label{fig:mnist}
 \end{figure}

 \begin{figure}[t]
   \captionsetup[subfigure]{labelformat=empty}
   \tiny
   \centering
   \includegraphics[width=0.5\linewidth]{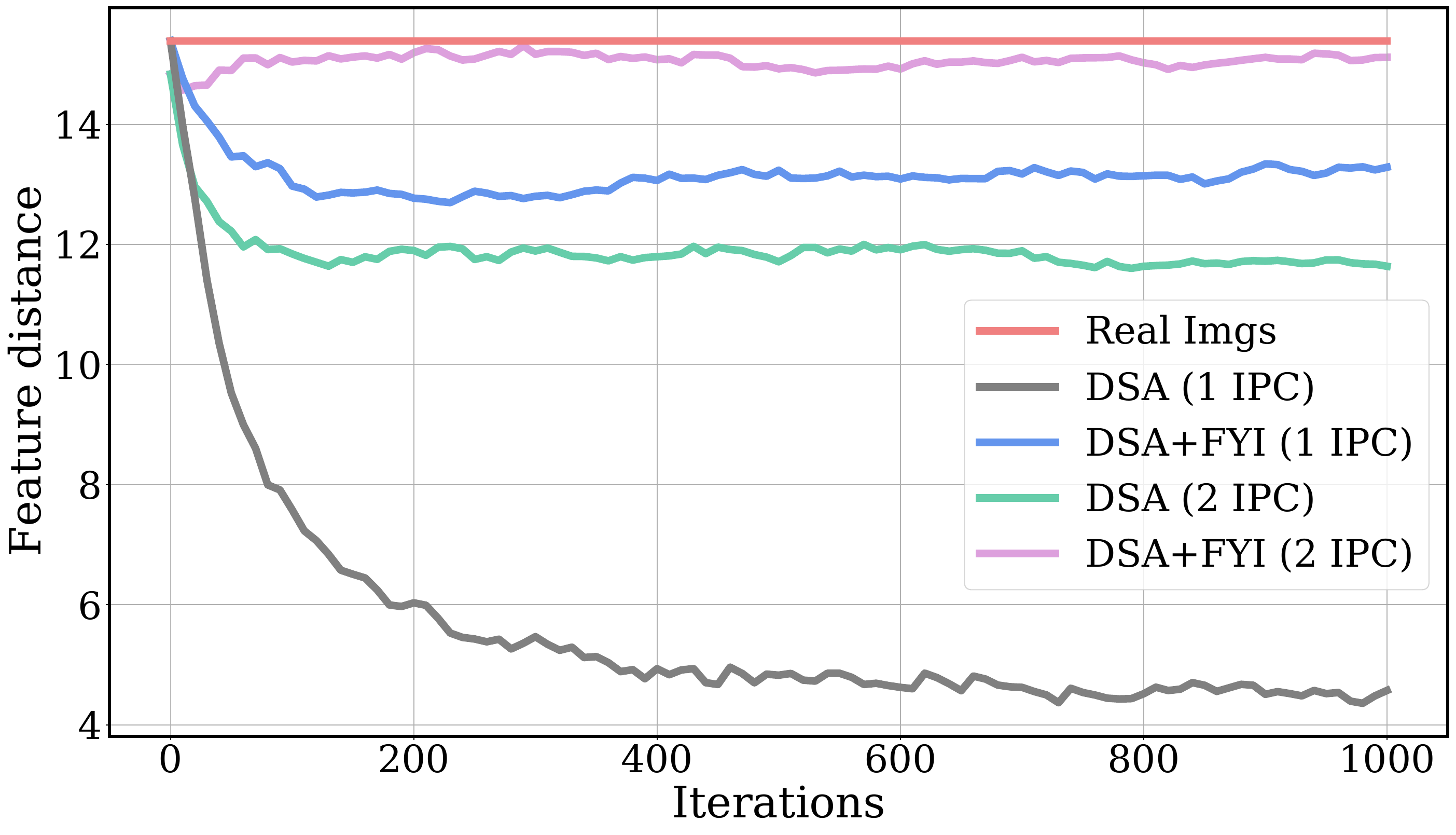}
   \vspace{-1em}
   \caption{The average distances between synthetic images and corresponding flipped counterparts in a feature space on the extended MNIST~\cite{mnist} dataset. We use ConvNet~\cite{conv} pre-trained on the real dataset to embed images into the feature space.}
   \label{fig:featdist}
   \vspace{-.5cm}
 \end{figure}

\subsubsection{Data augmentation.} We compare in Table~\ref{tab:ablation} the top-1 accuracy of FYI and its variants with different data augmentation techniques. Specifically, we rotate images by 15 degrees, scale them by a factor of 1.2, or flip them vertically followed by a batch-wise concatenation. We can see that 1) FYI outperforms all the variants, and 2) the variants mostly degrade or marginally improve the performance of the vanilla methods. To analyze the reason behind this result, we show in Fig.~\ref{fig:score_variants} the unequalness score and its variants on real images to further verify our interpretation. In detail, we replace the horizontal flipping in~\cref{eq:4} with other augmentation techniques and measure the scores with varying numbers of real images. We can see that the unequalness score converges to zero, while the variants do not. This is because these augmentation techniques other than horizontal flipping can generate samples that are out of the distribution of the original dataset. For example, as most of the objects are upright in images, synthesized images, if vertically flipped, correspond to samples from out of the distribution. Augmenting synthetic images using such techniques can prevent them from learning the semantics of the original dataset effectively. Note that real datasets are likely to be invariant under horizontal flipping. That is, similar patterns are highly likely to appear in different horizontal directions within a dataset, indicating that augmented samples from horizontal flipping belong to the in-distribution of the original dataset.

\subsubsection{Bilateral equivalence.} To further verify that FYI removes duplicated patterns and encodes rich semantics, we perform experiments with a dataset satisfying a perfect bilateral equivalence. Specifically, we apply a horizontal flipping to all images of MNIST~\cite{mnist} and construct an extended version consisting of an equal number of original and flipped images. We adopt DSA~\cite{dsa} to distill the augmented dataset into synthetic images. We can see in \cref{fig:mnist}(a) that the synthetic image from DSA is highly symmetric, making it difficult to recognize digits. In particular, the synthesized images of `3' and `8' become very similar, since DSA enforces the synthetic image of `3' to imitate both original and flipped images of `3'. On the contrary, we show in \cref{fig:mnist}(b) that DSA with FYI encodes different semantics on the left and right sides of images, leading to recognizable digits. Additionally, we show in \cref{fig:mnist}(c) the synthetic images using DSA for a 2 IPC case. Although we have more synthetic images, compared to \cref{fig:mnist}(b), the synthesized images are still symmetric (\eg, the number `2'). This implies that existing methods struggle to handle the bilateral equivalence, even with more number of synthetic images. Also, two images with the same class look very similar except for the directions, whereas those synthesized using FYI in \cref{fig:mnist}(d) look different in shapes. This indicates that our method further encodes rich semantics with more storage. We plot in Fig.~\ref{fig:featdist} the average Euclidean distance between an image and its flipped counterpart in a feature space during training. We can see that DSA using FYI preserves the distances comparable to those of the real images during training, suggesting that FYI mitigates the negative effects of bilateral equivalence, especially for the 1 IPC setting. On the contrary, the feature distances using DSA only decrease rapidly, indicating that both sides of synthetic images tend to contain duplicated patterns. The distance increases with the 2 IPC setting, as two images with the same class category are optimized together to capture different semantics. We can see that our method provides more asymmetric images even with the 1 IPC case, and its feature distances are almost the same as those for real images with the 2 IPC setting.

\subsubsection{Limitation.} Our method focuses on natural images that contain objects with arbitrary orientations, which could limit an application of our method to the dataset, where the orientation is important for recognition, typically containing numbers or characters.

\section{Conclusion}
We have presented a novel plug-and-play technique for dataset distillation, dubbed FYI, that enables better distilling rich semantics of real images into synthetic images. Specifically, we have found that object parts that appear on one side of a real image are highly likely to appear on the opposite side of another image within a dataset, making synthetic images of current methods fail to encode fine-grained details of objects. We have proposed a simple yet effective strategy that uses a horizontal flipping technique to encourage synthetic images to capture diverse information. Finally, we have shown that the proposed method can be easily integrated into state-of-the-art methods, demonstrating its effectiveness on standard benchmarks.

\subsubsection{Acknowledgements.} This work was partly supported by the NRF and IITP grants funded by the Korea government (MSIT) (No.2023R1A2C2004306, No.RS-2022-00143524, Development of Fundamental Technology and Integrated Solution for Next-Generation Automatic Artificial Intelligence System, No.2022-0-00124, Development of Artificial Intelligence Technology for Self-Improving Competency-Aware Learning Capabilities), and the Yonsei Signature Research Cluster Program of 2024 (2024-22-0161).

%
%
\bibliographystyle{splncs04}
\bibliography{main}



\section*{Supplementary material}

# FYI: Flip Your Images for Dataset Distillation Supplement

Byunggwan Son⬤, Youngmin Oh⬤,
Donghyeon Baek⬤, and Bumsub Ham⋆⬤

Yonsei University

In this supplementary material, we first provide more detailed explanations of existing dataset distillation methods (Sec. S1). We then describe more details for the experimental settings (Sec. S2). Finally, we present more analysis of our approach including results for continual learning and NAS (Sec. S3).

## S1 Current methods

We provide in the main paper a description of DM [19]. Here we review DC [18] and MTT [2] additionally.

**DC [18].** Several methods [7, 11, 12, 17], including DC, attempt to match one-step gradients through network parameters between real and synthetic datasets. Specifically, they use images for the same class to compute the distance metric, defined as follows:

$$D_\theta(\mathcal{T}_c, \mathcal{S}_c) = \sum_i \text{Cos}\Big[\mathcal{G}_i(\mathcal{S}_c;\theta), \mathcal{G}_i(\mathcal{T}_c;\theta)\Big], \qquad (i)$$

where Cos indicates the cosine distance between two vectors. We denote by $\mathcal{G}_i$ the vectorized gradients w.r.t. the parameters of the $i$-th output node. Note that an output node is defined as a filter in a convolutional layer and a vector to compute a neuron in a fully connected layer.

**MTT [2].** Extending the idea of matching one-step gradients, MTT matches training trajectories of real and synthetic datasets. Instead of minimizing a distance metric between images for the same class, MTT computes training trajectories using all images in the dataset and minimizes the distance between them. The overall objective function over different network parameters is defined as follows:

$$\mathcal{L} = \mathbb{E}_{\theta \sim P_\theta}\Big[D_\theta(\mathcal{T}, \mathcal{S})\Big], \qquad (ii)$$

where $\mathcal{T} = \bigcup_c \mathcal{T}_c$ and $\mathcal{S} = \bigcup_c \mathcal{S}_c$ are the real and synthetic datasets, respectively. In particular, MTT defines the distance metric $D_\theta$ as follows:

$$D_\theta(\mathcal{T}, \mathcal{S}) = \frac{\|\theta^\mathcal{S} - \theta^\mathcal{T}\|^2}{\|\theta - \theta^\mathcal{T}\|^2}, \qquad (iii)$$

---

⋆ Corresponding author.

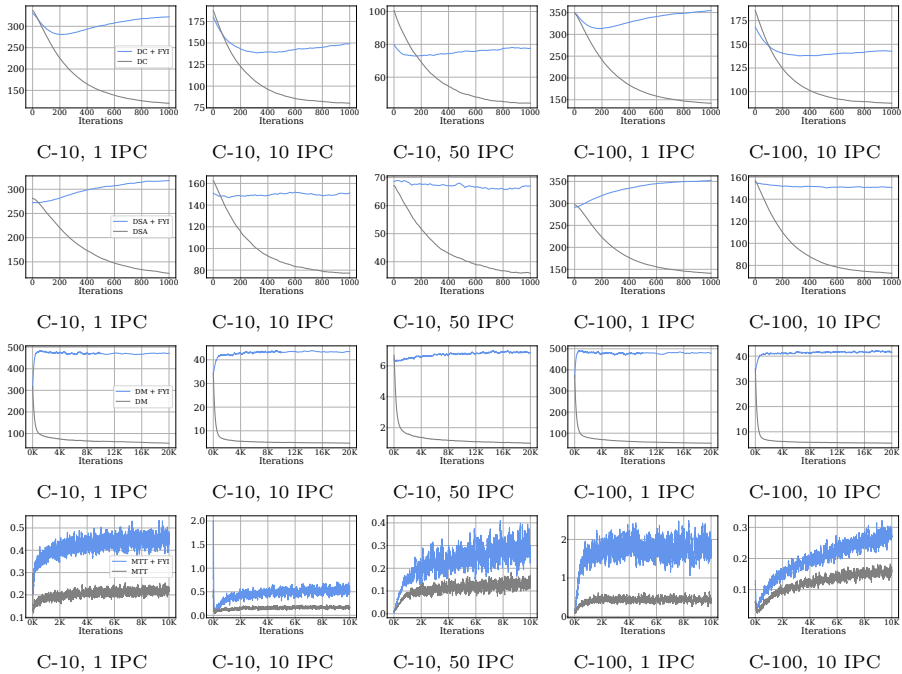

**Fig. A:** We show the unequalness scores of the synthesized datasets during training for DC [18], DSA [17], DM [19], and MTT [2] with and without FYI. We train synthetic datasets on CIFAR-10/100 [8] using 1, 10, or 50 IPC. We denote CIFAR-10 and CIFAR-100 by C-10 and C-100, respectively. FYI achieves higher unequalness scores throughout the training.

where $\theta^{\mathcal{T}}$ and $\theta^{\mathcal{S}}$ are network parameters updated from $\theta$ using $\mathcal{T}$ and $\mathcal{S}$ for a few iterations, respectively. Our FYI augments images in a mini-batch at each iteration to obtain the network parameters $\theta^{\mathcal{S}}$.

## S2   More details

**Hyperparameters.** Although FYI may achieve better results by adjusting various hyperparameters (*e.g.*, learning rate for images and networks), we follow the hyperparameters of the original papers. This shows that FYI does not require an additional search for hyperparameters. Exceptionally, we reproduce the results for MTT [2] and FTD [5] on Tiny-ImageNet [9] and FTD on CIFAR-100 under the 50 IPC setting using 2 times longer iterations to compute trajectories from synthetic images. We use efficient implementation of [4] to reduce memory caused by this change. This modification improves the performance regardless of the application of FYI.

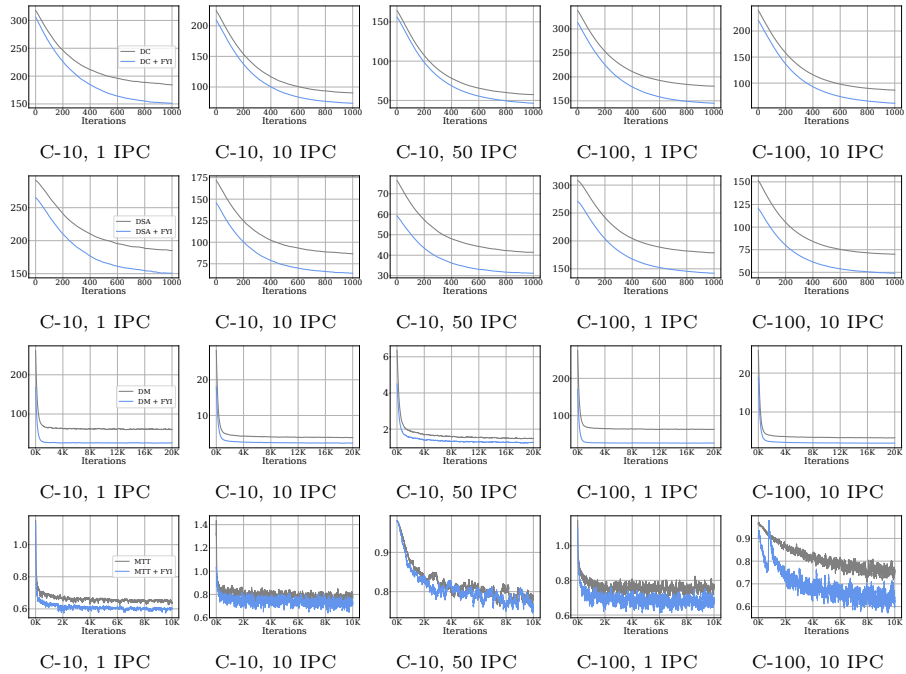

**Fig. B:** Plots of training losses computed using DC [18], DSA [17], DM [19], and MTT [2] with and without FYI. We train synthetic datasets on CIFAR-10/100 [8] with 1, 10, or 50 IPC settings. We denote CIFAR-10 and CIFAR-100 by C-10 and C-100, respectively. Using FYI lowers the training losses.

**Data preprocessing.** Data preprocessing technique used in the previous methods [2, 17–19] differs from each other. We use the same technique as the original papers to compare the performance with and without FYI. Specifically, we use standard channel-wise normalization for all the methods but also used ZCA transformation for MTT [2] for CIFAR-10 [8] with 1 and 10 IPC, and CIFAR-100 [8] with 1 and 50 IPC.

## S3 More results and discussions

**The unequalness score.** We show in Fig. A more results for the unequalness score using various methods [2, 17–19], compression ratios, and datasets [8]. We can see that FYI provides synthetic datasets with higher unequalness scores, and the score gaps are larger for higher compression ratios. This suggests the behavioral properties of the synthetic datasets and their flipped counterparts become more different using FYI and FYI encodes different semantics on different sides effectively.

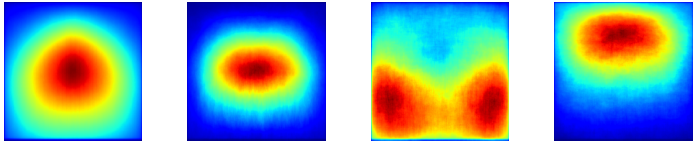

**Fig. C:** Distributions of objects on COCO. We show person, zebra, chair, and umbrella categories from left to right. Red: high, Blue: low.

| Dataset | CIFAR 10 | | | CIFAR 100 | |
|---|---|---|---|---|---|
| IPC | 1 | 10 | 50 | 1 | 10 |
| Ratio (%) | 0.02 | 0.2 | 1 | 0.2 | 2 |
| DC [18] | 28.3 (0.5) | 44.9 (0.5) | 53.9 (0.5) | 12.8 (0.3) | 25.2 (0.3) |
| DC + Flip | 27.9 (0.6) | 45.4 (0.5) | 54.0 (0.5) | 12.6 (0.4) | 27.5 (0.3) |
| DC + FYI | 30.0 (0.5) | 49.5 (0.5) | 54.6 (0.6) | 14.4 (0.3) | 29.2 (0.3) |
| DSA [17] | 28.8 (0.7) | 52.1 (0.5) | 60.6 (0.5) | 13.9 (0.3) | 32.3 (0.3) |
| DSA + Flip | 28.2 (0.6) | 52.0 (0.6) | 60.4 (0.4) | 14.1 (0.3) | 31.7 (0.3) |
| DSA + FYI | 30.6 (0.7) | 54.7 (0.5) | 63.7 (0.5) | 16.0 (0.3) | 35.0 (0.3) |
| DM [19] | 26.0 (0.8) | 48.9 (0.6) | 63.0 (0.4) | 11.4 (0.3) | 29.7 (0.3) |
| DM + Flip | 25.8 (0.6) | 49.3 (0.8) | 63.4 (0.4) | 11.1 (0.3) | 30.1 (0.3) |
| DM + FYI | 28.7 (1.2) | 53.1 (0.6) | 64.2 (0.4) | 13.3 (0.3) | 32.3 (0.4) |
| MTT [2] | 46.3 (0.8) | 65.3 (0.7) | 71.6 (0.2) | 24.3 (0.3) | 40.1 (0.4) |
| MTT + Flip | 47.0 (0.5) | 67.7 (0.4) | 72.8 (0.2) | 27.5 (0.7) | 41.1 (0.2) |
| MTT + FYI | 47.2 (0.4) | 68.2 (0.3) | 74.0 (0.3) | 28.2 (0.3) | 41.6 (0.2) |
| Full dataset | | 84.8 (0.1) | | | 56.2 (0.3) |

**Table A:** Quantitative comparison between applying random horizontal flipping for each synthetic image (Flip) and flip-and-concatenate strategy to all the synthetic images (FYI). We report the average top-1 accuracy (%) with standard deviations in the brackets on CIFAR-10/100 [8]. FYI is consistently better than Flip and Flip sometimes degrades the performance of the original methods.

**Training losses.** We plot in Fig. B the training losses during image synthesis over various settings. We observe that using FYI minimizes the training losses more effectively for all the settings. Minimizing the loss is achieved by capturing the rich semantics of real datasets into synthetic ones, and FYI achieves this by encoding diverse semantics across different sides in addition to different images.

**Distributions of objects.** We show in Sec. S2 distributions of locations for several objects on COCO. Each plot is obtained by counting how many times each location falls into a specific category using ground-truth masks. We can see that the distributions are also horizontally symmetric, implying that the bilateral equivalence exists in scene-centric datasets as well.

**Flipping without concatenation.** We compare in Tab. A our FYI, which flips and concatenates all the synthetic images, and the horizontal flipping technique for augmenting synthetic images during image synthesis. We can see that

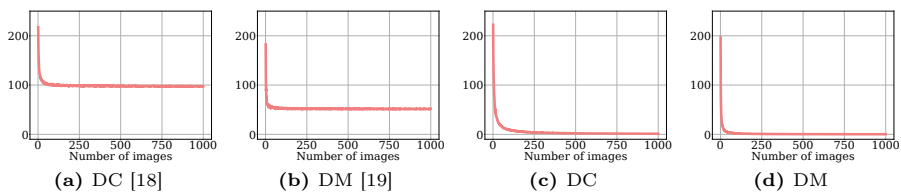

(a) DC [18]    (b) DM [19]    (c) DC    (d) DM

**Fig. D:** The bilateral equivalence measured using the unequalness score. We measure the unequalness score of real images sampled from MNIST [10] varying the number of samples. We use DC [18] and DM [19] to compute distance metrics. Before measuring the unequalness score, we apply horizontal flipping by 50% for each image in (c) and (d) whereas we do not apply any data augmentations for (a) and (b). The unequalness score consistently decreases when the bilateral equivalence is satisfied.

methods with FYI consistently outperform vanilla methods or those with horizontal flipping, confirming the effectiveness of our approach. This indicates that conditioning flipped images is important for dataset distillation. Note that the horizontal flipping does not improve vanilla methods in most of the settings. For MTT, where multiple updates are involved, using horizontal flipping consistently improves the performance as it would condition flipped images for updating synthetic images. Nonetheless, FYI still outperforms the horizontal flipping for MTT as well by conditioning flipped images with equal importance.

**The bilateral equivalence.** To show the effectiveness of the unequalness score as a measurement for the bilateral equivalence, we show in Fig. D (a, b) the unequalness score of real images sampled from MNIST [10] according to the number of images. We can see that the unequalness score does not decrease even when the number of images are increased for MNIST dataset since the images are aligned in a specific direction. On the contrary, when we apply random horizontal flipping before measuring the unequalness score, the score consistently decreases as the number of images increases Fig. D (c, d). It is noteworthy that we sample images without replacement and all the images are distinct. The results suggest that the unequalness score is a good measurement for the bilateral equivalence and the score decreases when similar patterns are observed across different sides within a group of images.

**Continual learning.** In class-incremental continual learning [3, 13, 16], images belonging to certain class categories are given at each training step and networks are evaluated to classify among all the class categories that are seen at the past and current training steps. To let networks predict class categories in the past steps accurately, many methods [1, 13, 14] store images in a memory buffer and use them to train networks together with the images from the current step. To test the effectiveness of dataset distillation in constructing memory buffers, previous works [15, 18, 19] store synthetic images in a memory buffer and train networks from scratch using the images in the memory buffer only. We note that

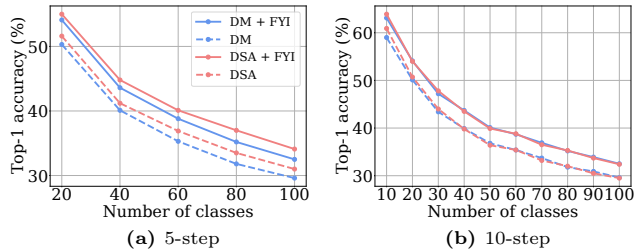

(a) 5-step  (b) 10-step

**Fig. E:** Quantitative comparison between current methods [17, 19] with and without FYI for continual learning on CIFAR-100 [8]. We plot the average top-1 accuracy (%) of the networks trained on 10 IPC synthesized in 5- and 10-step class-incremental settings.

|            | Top 50% | All   | Time (min) | Images No. |
|------------|---------|-------|------------|------------|
| Real       | 1.000   | 1.000 | 6,804      | 50,000     |
| MTT        | 0.312   | 0.670 | 360        | 500        |
| MTT + FYI  | **0.382** | **0.699** | 360   | 500        |

**Table B:** Quantitative comparison between MTT [2] with and without FYI for NAS on CIFAR-10 [8] with 50 IPC. We report Spearman's ranking correlation to the results from the whole dataset.

the size of the memory buffer grows linearly to the number of object categories to classify. We divide 100 classes of CIFAR-100 into 5 steps or 10 steps with the same number of classes for each step and synthesize 10 IPC, which belong to the class categories of the current step. We randomly select class categories without replacement, and train randomly initialized ConvNets [6]. We plot in Fig. E the average top-1 accuracy (%) of the networks across different random seeds using DM [19] and DSA [17], with and without FYI. From the results, we can see that FYI improves the performance of vanilla methods for both 5-step and 10-step class-incremental learning. We argue that encoding fine-grained details into the images in the memory buffer is important for continual learning, and FYI can provide such details.

**NAS.** As an application of dataset distillation, previous works [5, 18, 19] train various networks in a predefined search space using synthetic datasets and rank the architectures based on the validation accuracy. When the rankings align with those predicted from real datasets, synthetic datasets can act as a proxy for the real datasets for NAS reducing considerable computational burden and memory requirements. We construct a search space consisting of 720 ConvNets using different combinations of width, depth, normalization, activation function, and pooling operations, following the previous works. We compare in Table B Spearman's ranking correlation between the rankings of the synthetic and the real dataset using MTT [2] with and without FYI. Specifically, we synthesize

50 IPC on CIFAR-10 [8] and compare the correlation with top-performing 50% architectures searched by the real dataset or all the architectures. We can see that MTT with FYI provides higher correlations compared to the vanilla MTT in both settings. This demonstrates that FYI is effective in searching for architectures that perform well on real datasets because FYI incorporates diverse semantics into the synthetic datasets.



\end{document}